\documentclass{article} 
\usepackage{iclr2026_conference,times}

\usepackage{amsmath,amsfonts,bm}

\def\eqref#1{equation~\ref{#1}}

\def\1{\bm{1}}

\def\vh{{\bm{h}}}

\def\vp{{\bm{p}}}

\def\vw{{\bm{w}}}
\def\vx{{\bm{x}}}

\def\vz{{\bm{z}}}

\def\mE{{\bm{E}}}

\def\mW{{\bm{W}}}

\DeclareMathAlphabet{\mathsfit}{\encodingdefault}{\sfdefault}{m}{sl}
\SetMathAlphabet{\mathsfit}{bold}{\encodingdefault}{\sfdefault}{bx}{n}

\usepackage{amsmath,amssymb,amsfonts,amsthm}
\usepackage{graphicx}
\usepackage{subcaption}
\usepackage{booktabs}
\usepackage{threeparttable}
\usepackage{microtype}
\usepackage{multirow}
\usepackage{hyperref}
\usepackage{url}

\graphicspath{{./figures/}}

\title{OCELOT: Planetary-Scale Species Distribution Modeling with Satellite Foundation Embeddings}

\author{Bart van Merriënboer \\
Google DeepMind \\
\texttt{bartvm@google.com} \\
\And Lauren Harrell \\
Google Research \\
\And Keith Anderson \\
Google DeepMind \\
\And Burcu Karagol Ayan \\
Google DeepMind \\
\And Elise Kleeman \\
Google Research \\
\And Tom Denton \\
Google DeepMind \\
\And Drew Purves \\
Google DeepMind \\
}

\iclrfinalcopy 
\begin{document}

\maketitle
\lhead{}
\renewcommand{\headrulewidth}{0pt}

\begin{abstract}
Large-scale repositories of species observations now contain billions of occurrence records from around the world, but these sightings cluster heavily in human-accessible areas. Existing models use only a small fraction of these records and are not evaluated on extrapolating to remote, unvisited regions---where species distribution models can be used as geographic priors to help classify data from camera traps and acoustic recorders. We scale relative occurrence species distribution models by training a multimodal implicit neural representation network called OCELOT that fuses spatial coordinates with satellite foundation embeddings on two orders of magnitude more observations than previous work: 2 billion GBIF records across 100,000 taxa. Scaling to this regime requires co-designing the data pipeline, taxonomic representation, loss formulation, and model architecture: spatiotemporal thinning and sub-linear cell sampling correct for spatial redundancy and observer bias, while an additive taxonomic readout shares representations across species, genera, and families to improve predictions for rare taxa. To evaluate performance in remote areas, we benchmark geographic extrapolation using the NDCG metric on spatial block holdouts up to 600\,km. Satellite foundation embeddings improve extrapolation by 51\% over coordinate-only models and outperform classical bioclimatic rasters. The resulting model also matches or outperforms existing baselines on established interpolation benchmarks, demonstrating that optimizing for geographic extrapolation does not compromise in-distribution accuracy.
\end{abstract}

\section{Introduction}
\label{sec:intro}

Understanding the geographic distributions of biological species is essential for monitoring biodiversity loss, designating protected areas, and forecasting ecological responses to climate change \citep{elith2009species}. Biodiversity databases such as the Global Biodiversity Information Facility (GBIF) and iNaturalist now contain billions of occurrence records across hundreds of thousands of species, and recent neural approaches such as SINR \citep{cole2023spatial} have demonstrated that deep networks can model tens of thousands of species jointly. However, existing neural SDMs and their evaluations focus on spatial interpolation: test queries are drawn from the same human-accessible locations that dominate the training data. Conservation and ecological monitoring increasingly rely on autonomous sensors---such as camera traps and passive acoustic recorders---deployed in remote areas \citep{zbinden2024selection}. Because SDMs serve as geographic priors to help classify detections from these sensors, models must generalize to regions with few or no prior sightings: they must \emph{extrapolate}, not interpolate.

Opportunistic occurrence records are presence-only: absences are unrecorded. Presence-only data identifies \emph{relative occurrence}: the probability that an observation at a given location belongs to one species rather than another, reflecting local community composition (\S\ref{sec:problem_formulation}). Relative occurrence is directly applicable as a geographic prior for automated species identification \citep{macaodha2019presence,cole2023spatial}: when a camera trap or acoustic recorder captures an observation, the SDM provides a spatial prior over which species are likely at that location. Note that this estimand is distinct from absolute occupancy or range boundaries---a distinction that prior neural SDMs such as SINR \citep{cole2023spatial} blur by treating presence-only records as binary occupancy labels.

Scaling relative occurrence modeling to 100,000 taxa and 2 billion records---two orders of magnitude more observations than previous neural SDMs---while ensuring that the model is suitable for deployment in remote areas presents several challenges. First, the model must \emph{extrapolate geographically}. For extrapolation, the model needs environmental features that generalize across space, and a training distribution corrected for the covariate shift between human-accessible areas and remote areas. Prior work concluded that environmental covariates provide negligible benefit over spatial coordinates alone \citep{cole2023spatial}; however, that conclusion was drawn from interpolation-only evaluation. Second, the model must \emph{handle the taxonomic long tail}: most species have few observations, and the model must share statistical strength across related taxa without sacrificing predictions for common species. Third, the \emph{data pipeline must scale}. Billions of occurrence records are spatially redundant and concentrated in biodiversity hotspots; the pipeline must deduplicate sightings, aggregate records into spatial cells, and account for uneven observer effort across locations while stabilizing training despite the heavy-tailed distribution of per-cell observation counts. Fourth, extrapolation performance must be \emph{explicitly evaluated}. Standard interpolation benchmarks cannot distinguish a model that memorizes the observer footprint from one that generalizes; the evaluation protocol must test prediction using spatial block holdouts. We call our model OCELOT: \textbf{O}ccurrence \& \textbf{C}oordinate \textbf{E}xtrapolation for \textbf{L}ong-tail \textbf{O}rganism \textbf{T}axa.

Our contributions address these four challenges:
\begin{enumerate}
    \item We train a multimodal implicit neural representation that fuses spatial coordinates with satellite foundation embeddings (AEF; \citealt{brown2025alphaearth}). To correct for the covariate shift between the training distribution and remote deployment areas, we down-weight heavily observed grid cells during training so that sparsely observed regions receive more weight. AEF improves extrapolation by 51\% over coordinate-only models and effectively supersedes classical bioclimatic rasters.
    \item We handle the taxonomic long tail of 100,000 species with an additive taxonomic readout that shares representations across species, genera, and families, combined with independent sigmoid losses and class-prior offsets, demonstrating gains on data-scarce rare species.
    \item We design a data pipeline that compresses 2.9 billion raw records into spatial cells using spatiotemporal thinning, multi-label cell aggregation, and target-group background conditioning to cancel shared observer effort. Rather than reweighting the loss---which causes high gradient variance when hotspot cells contain hundreds of thousands of sightings---we sample cells proportional to their observation count, keeping gradients stable while compressing the dataset 15$\times$.
    \item We evaluate geographic extrapolation using Normalized Discounted Cumulative Gain (NDCG) on spatial block holdouts up to 600\,km alongside standard interpolation benchmarks (geo-prior accuracy, eBird S\&T, and IUCN range maps), achieving strong extrapolation while matching or outperforming prior models on interpolation.
\end{enumerate}

\section{Related Work}
\label{sec:related_work}

\subsection{Ecological Modeling and Point Processes}
Classical species distribution modeling relies on Maximum Entropy (MaxEnt; \citealt{phillips2006maximum}), Generalized Additive Models (GAMs; \citealt{elith2009species}), and Inhomogeneous Poisson Processes (IPPs; \citealt{warton2010poisson,renner2015point}). These methods account for sampling effort by contrasting presence records against target-group background sightings \citep{phillips2009sample,fithian2015bias,barber2021target}. They usually fit independent single-species models against coarse features (e.g., a handful of bioclimatic variables; \citealt{fick2017worldclim}), with no mechanism to share representations across species that share habitats. Joint Species Distribution Models (JSDMs; \citealt{kissling2012towards,pollock2014understanding,warton2015so,wilkinson2021defining}) were designed to capture species co-occurrences via covariance matrices or latent factor models. However, estimating pairwise interaction matrices scales quadratically in the number of species, restricting most JSDMs to small communities.

\subsection{Implicit Neural Representation Models}
Recent work has applied deep networks to spatial ecology \citep{botella2018deep,macaodha2019presence}. SINR \citep{cole2023spatial} learns continuous species ranges from geographic coordinates, and \citet{russwurm2024geographic} showed that spherical harmonic basis functions provide a better spatial basis than Fourier features for location encoding. These models performed strongly on interpolation tasks but not were evaluated for extrapolation (e.g., using spatial block holdouts). Additionally, these models treated species as flat, unstructured classes, ignoring phylogenetic relationships.

\subsection{Earth Observation Foundation Models and Long-Tailed Learning}
Earth observation foundation models such as AlphaEarth Foundations \citep{brown2025alphaearth} provide dense, multi-spectral embeddings that capture vegetation structure, canopy phenology, and surface hydrology globally. Separately, the long-tailed recognition literature has explored class-prior calibration \citep{menon2021longtail} and independent sigmoid losses \citep{zhai2023sigmoid} to train classifiers under extreme class imbalance without softmax competition. We draw on both lines of work: we fuse satellite foundation embeddings with spherical harmonic spatial encodings through a two-tower architecture, and we use an additive taxonomic readout with independent sigmoid losses to handle the taxonomic long tail.

\section{Methodology}
\label{sec:methodology}

\subsection{Problem Formulation: Relative Occurrence vs.\ Multi-Label Occupancy}
\label{sec:problem_formulation}

Presence-only records cannot identify absolute species occupancy because non-detection is unrecorded \citep{fithian2013finite,mackenzie2002estimating}. Let $x \in \mathcal{X}$ denote an observation location (encoding both geographic coordinates and local environmental covariates). Sighting events for species $s \in \{1, \dots, S\}$ are governed by an Inhomogeneous Poisson Process (IPP) \citep{warton2010poisson} with rate parameter:
\begin{equation}
    \lambda_s(x) = \alpha_s(x) \cdot e_s(x),
\end{equation}
where $\alpha_s(x)$ is the ecological occurrence intensity and $e_s(x)$ is the observation effort for species $s$---encompassing site-level sampling intensity (how often a location is visited) and species-specific detection probability (how conspicuous or identifiable the taxon is). Because absences are unrecorded, $e_s(x)$ and $\alpha_s(x)$ are confounded: zero sightings cannot distinguish unsuitable habitat ($\alpha_s(x) = 0$) from unvisited areas ($e_s(x) = 0$).

Target-group background (TGB) methods \citep{phillips2009sample,fithian2013finite,barber2021target} address this confound under the assumption that detection effort is approximately shared across species: $e_s(x) \approx e(x)$ for all $s$. This proportional-bias assumption is standard in presence-only SDMs \citep{phillips2009sample,fithian2015bias}---it is an approximation (some taxa are more conspicuous than others), but the best available when species-specific observation effort cannot be estimated. Under this assumption, conditioning on the union of all sightings cancels the shared effort:
\begin{equation}
    P\big(Y = s \;\big|\; X = x, \; \text{observation at } x\big) = \frac{\alpha_s(x) e(x)}{\sum_{s'=1}^S \alpha_{s'}(x) e(x)} = \frac{\alpha_s(x)}{\sum_{s'=1}^S \alpha_{s'}(x)}.
    \label{eq:relative_occurrence}
\end{equation}
Presence-only SDMs trained with TGB therefore estimate \textbf{relative occurrence}: given that an organism was documented at $x$, what fraction of records belongs to species $s$? This reflects community composition rather than absolute occupancy (see Appendix~\ref{app:tgb_nce} for the residual when proportional bias fails).

Note that some prior work~\citep{cole2023spatial} formalizes presence-only records as multi-label binary occupancy ($Y_s \in \{0, 1\}$). This treats unobserved taxa as negative absences, distorting probability calibration in species-rich communities where relative encounter rates are necessarily small even for present species.

A common application of relative occurrence is to serve as a geographic prior for automated species identification \citep{macaodha2019presence,cole2023spatial}, where geographic predictions modulate the output of a vision or audio classifier. Assuming conditional independence of the sensor input $I$ (e.g., a camera trap image or audio recording) and location $x$ given species ($I \perp x \mid s$), Bayes' rule yields:
\begin{equation}
    P(s \mid I, x) \propto P(s \mid I) \cdot P(s \mid \text{observation at } x) \,/\, P(s),
\end{equation}
where $P(s \mid I)$ is the classifier output, $P(s \mid \text{observation at } x)$ is the relative occurrence estimated by our model (Equation~\ref{eq:relative_occurrence}), and dividing by the species base rate $P(s)$ prevents double-counting the marginal class prior.

\subsection{Data Pipeline: Thinning, Aggregation, and Spatial Sampling}
\label{sec:thinning_aggregation}

Biodiversity databases can contain substantial redundant sightings: observers logging multiple photographs of the same organism during a single visit, or multiple observers converging to report the same localized individual (e.g., a rare bird sighting in a popular park). Left untreated, these duplicate records inflate the empirical risk on heavily visited sites and taxa. For this reason, spatial thinning is standard practice in SDMs~\citep{aiellolammens2015spthin}.

\begin{figure}[t]
\centering
\begin{minipage}[t]{0.48\linewidth}
  \vspace{0pt}
  \centering
  \includegraphics[width=\linewidth]{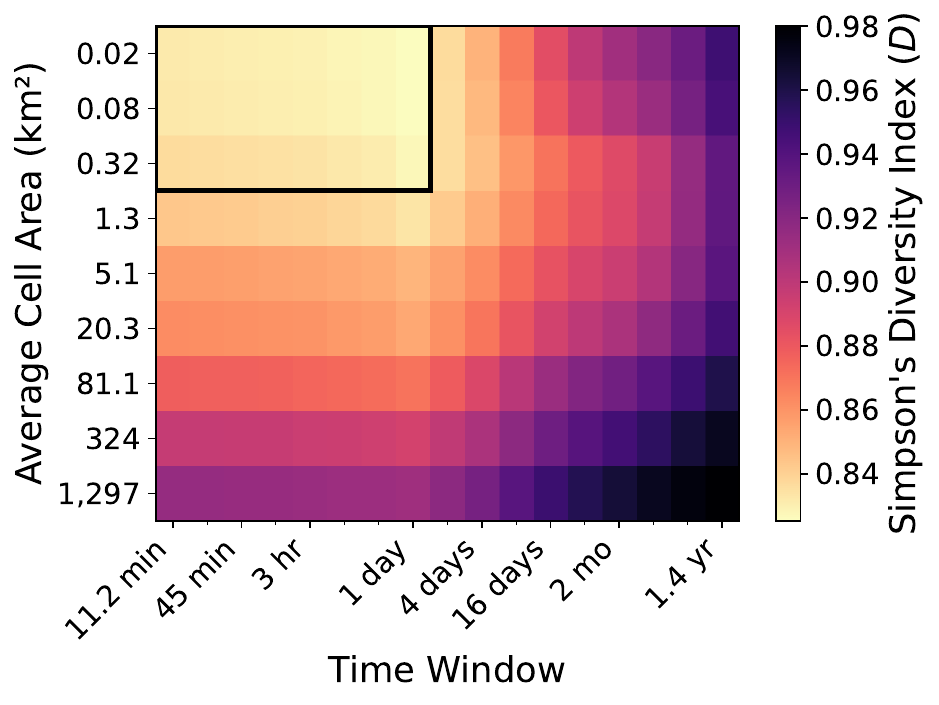}
  \caption{\textbf{Spatio-temporal Simpson's diversity across space and time.} Evaluated across GBIF occurrences. The black box marks our chosen thinning window ($\le 1$ day, S2 level 14, $\sim 600$\,m), capturing intra-visit duplicates at the empirical noise floor before background community turnover.}
  \label{fig:spatiotemporal_thinning}
\end{minipage}\hfill
\begin{minipage}[t]{0.48\linewidth}
  \vspace{0pt}
  \centering
  \includegraphics[width=\linewidth]{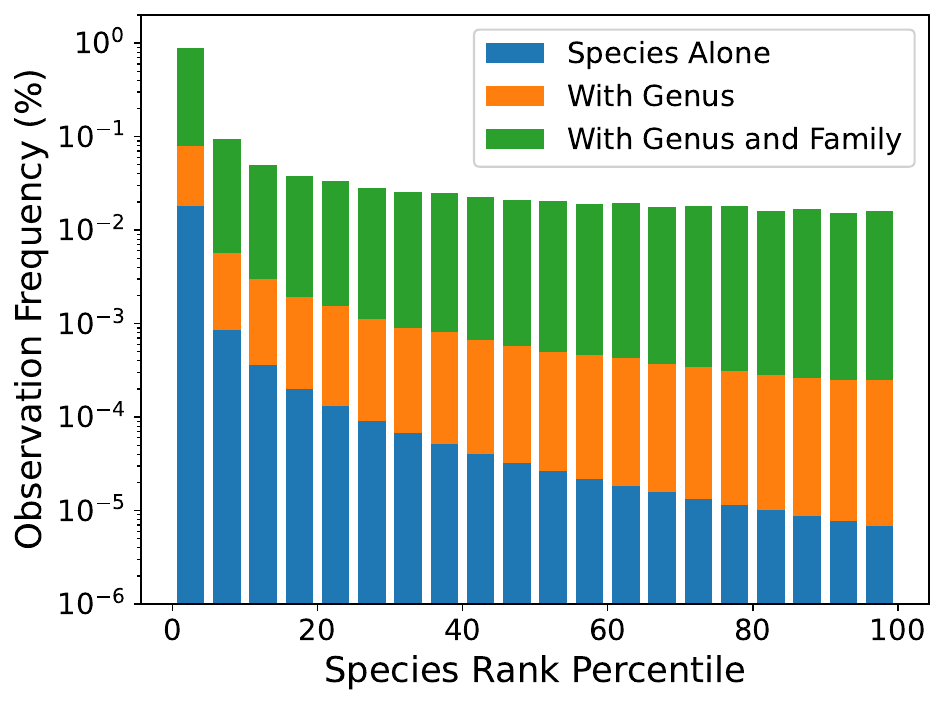}
  \caption{\textbf{Taxonomic update frequency gain across the long tail.} Species partitioned into 20 rank bins of 5,000 species each. Stacked bars show additive progression from species ($\mE_{\text{species}}$, blue), to genus ($\mE_{\text{genus}}$, orange), to family ($\mE_{\text{family}}$, green).}
  \label{fig:binned_update_frequency}
\end{minipage}
\end{figure}

To select thinning cutoffs objectively rather than through arbitrary heuristics \citep{steen2021evaluation}, we analyze the \textbf{spatio-temporal Simpson's diversity index}. Let each occurrence record $i$ be defined by its spatial location $x_i$, timestamp $t_i$, and observed species identity $y_i$. For a spatial cell $c$ at S2 grid resolution $\text{level}$ and a temporal window of duration $\Delta t$, we evaluate the probability that two proximate sightings belong to different species:
\begin{equation}
    D(\text{level}, \Delta t) \equiv P\Big(y_i \ne y_j \;\Big|\; x_i, x_j \in c, \; |t_i - t_j| \le \Delta t\Big).
\end{equation}
This formulation extends Shimatani's \citeyearpar{shimatani2001multivariate} spatial Simpson index to space-time windows. Because organisms cluster locally in space and exhibit movement and migration across time, diversity naturally decreases as the space-time window shrinks. As shown in Figure~\ref{fig:spatiotemporal_thinning}, diversity drops to a minimum plateau ($\approx 0.83$) for single-day visits ($\Delta t \le 1$\,day). Spatially, diversity reaches its minimum at S2 level 14 ($\approx 600$\,m diameter). For smaller windows, location uncertainty and organism movement scatter repeat encounters across neighboring cells, causing measured diversity to flatten. We therefore thin the dataset at this empirical noise floor by retaining at most one record per species per day within each S2 level 14 cell, eliminating 28.8\% of records.

Following spatio-temporal thinning at S2 level 14 ($\approx 600$\,m), co-located occurrences are aggregated into multi-label cells at S2 level 20 ($\approx 10$\,m, matching the native pixel grid of the satellite foundation embeddings). For each spatial cell $i$, we construct a normalized multi-label target vector $\vp_i = (c_{i,1}/N_i, \dots, c_{i,S}/N_i)$, where $c_{i,s}$ is the count of species $s$ and $N_i = \sum_s c_{i,s}$. Combined with thinning, this reduces raw records $15\times$ ($10.5\times$ from cell aggregation alone), significantly speeding up training.

Keeping the expected gradient equivalent to training on raw un-aggregated records requires scaling each cell's loss by its observation count $N_i$. However, observation counts follow a heavy-tailed Pareto distribution with tail index $\hat{\alpha} \approx 1.0 < 2$ (Appendix~\ref{app:subsec_pareto_tail}): the second moment $\mathbb{E}[N_i^2]$ diverges, so loss weighting leads to high gradient variance and unstable training \citep{ionides2008truncated,vehtari2024pareto}. Instead, we draw cells stochastically with probability proportional to $N_i$ via Vose's alias method, which preserves the expected gradient:
\begin{equation}
    \mathbb{E}_{i \sim P}[\nabla \mathcal{L}_i] = \sum_i \frac{N_i}{\sum_m N_m} \sum_{s \in \mathcal{P}_i} \frac{c_{i,s}}{N_i} \nabla \ell_{i,s} = \frac{1}{\sum_m N_m} \sum_i \sum_{s \in \mathcal{P}_i} c_{i,s} \nabla \ell_{i,s}.
\end{equation}

Citizen-science data is heavily biased toward accessible, frequently visited areas \citep{kramerschadt2013importance}, creating a covariate shift relative to remote deployment regions \citep{shimodaira2000improving}. Because our spatial block holdout evaluation assumes a uniform geographic prior (Section~\ref{sec:eval_metrics}), correcting for this shift requires down-weighting heavily observed areas. Our resampling framework provides a natural mechanism: we introduce a temperature $T_{\text{cell}} \in [0, 1]$ that interpolates between the record-level distribution and a uniform location-level distribution:
\begin{equation}
    w_i \propto (N_i)^{T_{\text{cell}}}.
\end{equation}
Setting $T_{\text{cell}} = 1.0$ recovers the record-level objective but overweights heavily visited observation hotspots, while $T_{\text{cell}} = 0.0$ weights locations uniformly, allowing noisy single-observation cells to dominate. We set $T_{\text{cell}} = 0.5$ based on hyperparameter optimization (Section~\ref{sec:ablations}), which shifts the training objective toward the uniform evaluation prior while retaining a preference for cells whose multi-label vectors are more reliable.

\subsection{Model Architecture}
\label{sec:architecture}

OCELOT predicts species occurrence from a queried location $x$, represented by two feature vectors: a $(L+1)^2$-dimensional spatial encoding ($\mathbf{v}_{\text{spatial}}$) derived from spherical harmonics, and a 64-dimensional environmental embedding ($\mathbf{v}_{\text{env}}$) provided by AlphaEarth Foundations (AEF) satellite representations \citep{brown2025alphaearth}. High-degree spherical harmonics allow the model to learn continuous spatial frequency patterns without boundary artifacts. Concurrently, the AEF embeddings capture a wide variety of signals---topography, phenology, canopy structure, and human footprint---providing the essential environmental context needed to extrapolate predictions into unvisited regions.

Each feature modality is first processed by an independent 2-block residual tower to project into a latent space $\mathbb{R}^d$ ($d=512$). The outputs of these independent towers are then concatenated and processed by a shared deep Pre-LayerNorm residual MLP trunk \citep{wang2020makes} of $N=4$ residual blocks:
\begin{equation}
    \vz^{(l+1)} = \vz^{(l)} + \mW_2^{(l)} \, \text{Dropout}\Big(\text{GELU}\big(\mW_1^{(l)} \, \text{LayerNorm}(\vz^{(l)})\big)\Big),
\end{equation}
producing the final location embedding $\vh = \vz^{(N)} \in \mathbb{R}^d$.

\paragraph{Hierarchical taxonomic readout.}
\label{sec:hierarchical_readout}
Rather than learning an independent weight vector $\vw_s$ per species, we decompose it into a sum of per-rank embeddings. Let $r_k(s)$ denote the taxon index of species $s$ at rank $k \in \{1, \dots, K\}$ (species, genus, and family). The readout computes:
\begin{equation}
    \vw_s = \sum_{k=1}^K \mE_k[r_k(s)],
    \label{eq:hierarchical_readout}
\end{equation}
where $\mE_k \in \mathbb{R}^{N_k \times d}$ is a learned embedding table for rank $k$ ($N_1 = 100{,}000$ species, $N_2 = 29{,}684$ genera, $N_3 = 4{,}484$ families). We truncate the hierarchy at the family rank ($K=3$, omitting order, class, phylum, and kingdom) because phylogenetic niche conservatism operates robustly across genera and families \citep{hadly2009niche,crisp2009phylogenetic}, where shared physiological and morphological constraints maintain broad biome affinities \citep{wiens2005niche}. We find that going beyond $K=3$ yields diminishing returns (Section~\ref{sec:ablations}).

With a flat readout, species in the bottom 5\% (mean 139 occurrences) account for fewer than $6.8 \times 10^{-6}\%$ of training examples, leaving their weight vectors effectively untrained. Under the additive readout, whenever \textit{any} sister taxon in genus $g(s)$ or family $f(s)$ appears, gradients update $\mE_{\text{genus}}[g(s)]$ and $\mE_{\text{family}}[f(s)]$. As shown in Figure~\ref{fig:binned_update_frequency}, the ratio of parent-rank to species-level occurrences reaches $\mathbf{33.8\times}$ (genus) and $\mathbf{2{,}344\times}$ (family) for the rarest 5\% of species. This allows rare species to rely on well-trained genus and family representations, while abundant species learn species-specific adjustments on top of the shared hierarchy.

\subsection{Loss Formulation}
\label{sec:loss_formulation}
\label{sec:optimization_dynamics}

To account for extreme class imbalance across 100,000 species, species logits incorporate a fixed class-prior offset:
\begin{equation}
    z_s = \vh^T \vw_s + \log \frac{\pi_s}{1 - \pi_s},
    \label{eq:class_prior}
\end{equation}
where $\pi_s = N_s / \sum_m N_m$ is the empirical marginal frequency (prevalence). This anchors each species' baseline prediction to its global catalog abundance, allowing the network $\vh^T \vw_s$ to focus parameter capacity on learning local geographic and environmental deviations from the prior.

To optimize these logits, standard softmax cross-entropy is problematic: the most frequent species (20.9M occurrences) dominate the partition function, persistently pushing tail species logits downwards \citep{tan2020equalization,kang2020decoupling}. We remove this inter-species gradient competition by replacing the softmax with $S$ independent binary cross-entropy (BCE) sigmoid losses \citep{zhai2023sigmoid}:
\begin{equation}
    \mathcal{L}_{\text{BCE}} = -\sum_{s=1}^S \Big[ y_s \log \sigma(z_s) + (1 - y_s) \log(1 - \sigma(z_s)) \Big].
    \label{eq:bce_loss}
\end{equation}
The negative gradient on class $j$ is now $\sigma(z_j)$ (for $y_j = 0$), which depends only on $z_j$ and not on the logits of other classes. By noise-contrastive estimation (NCE) theory \citep{gutmann2010noise,mnih2012fast}, unnormalized logistic classifiers trained against background samples converge to the true conditional density. In practice, the total probability deviates slightly from one, so we normalize the sigmoid outputs at test time, recovering proper conditional probabilities $P(Y = s \mid X = x)$. Note that each species' binary classifier receives positives from its own observations and negatives from all other species' observations---the same training data as a target-group background (TGB) classifier \citep{phillips2009sample}. Our multi-label formulation is therefore equivalent to fitting $S$ independent TGB models; Appendix~\ref{app:tgb_nce} provides the full derivation.

Because each spatial cell contains only a sparse subset of observed species $\mathcal{P}_i \subset \{1, \dots, S\}$ ($|\mathcal{P}_i| \ll S$), we store labels as sparse arrays and rewrite Equation~\ref{eq:bce_loss} as an unconditional softplus reduction plus a sparse positive correction over the cell composition vector $\vp_i$:
\begin{equation}
    \mathcal{L}_i = \sum_{s=1}^S \text{softplus}(z_{i,s}) - \sum_{s \in \mathcal{P}_i} p_{i,s} z_{i,s},
\end{equation}
avoiding the need to materialize intermediary dense $B \times S$ label matrices (for batch size $B$) and enabling training on a single GPU.

\section{Experiments and Results}
\label{sec:experiments}

\subsection{Experimental Setup}
\label{sec:experimental_setup}

Our training data is derived from the September 2026 occurrence snapshot of GBIF. We remove records that GBIF identifies as having quality issues (e.g., invalid coordinates) and match the remaining records to an active species in the iNaturalist taxonomy using the scientific name, discarding records that cannot be matched. The resulting dataset comprises 2.91 billion records across 480,013 species. Spatiotemporal thinning at S2 level 14 ($\approx 600$\,m resolution) per species-day-cell removes 838.7 million redundant sightings (28.8\%), leaving 2.07 billion thinned occurrences across 29.7 million unique S2 level-14 cells. Co-located records are aggregated into multi-label cells at S2 level 20 ($\approx 10$\,m), yielding 198.1 million spatial training cells. Our vocabulary consists of the top $S = 100{,}000$ species, covering 99.66\% of all thinned occurrences. Occurrence frequencies span five orders of magnitude: from 20.9 million for the most common species to 600 at rank 50,000, and 132 at rank 100,000. Species belong to 29,684 genera and 4,484 families.

For each occurrence record, we extract the corresponding spatial and environmental descriptors:
\begin{itemize}
    \item \textbf{Spherical harmonics:} Geographic coordinates are expanded up to degree $L=50$, producing precisely $2{,}601$ spatial components. While higher degrees exist, we empirically fix $L=50$ to balance detail against the risk of memorizing local observer density (see degree ablations in Section~\ref{sec:ablations}).
    \item \textbf{AlphaEarth Foundations (AEF):} Features are extracted via Google Earth Engine \citep{brown2025alphaearth}, using the nearest year available for GBIF occurrences and the year 2022 for the interpolation tasks. Although queried at 10\,m grid spacing, the underlying architecture integrates visual context over a $1.28 \times 1.28$\,km receptive field (see Appendix~\ref{app:aef_spatial_resolution} for an empirical variogram analysis).
    \item \textbf{WorldClim BioClim:} To benchmark against classical species distribution models, we additionally extract the 19 standard bioclimatic variables \citep{fick2017worldclim} at 30-arcsecond ($\approx 1$\,km) resolution.
\end{itemize}

Models are implemented in JAX and trained for 5M steps ($\approx 20$\,hours at $70$\,steps/s) with AdamW on a single NVIDIA H100 GPU with batch size 4,096. We use a cosine learning rate schedule with peak learning rate $3 \times 10^{-4}$, 10,000 warmup steps, and weight decay $3 \times 10^{-4}$.

\subsection{Evaluation Protocol: Interpolation and Extrapolation}
\label{sec:eval_metrics}

To prevent spatial data leakage from autocorrelation \citep{roberts2017crossvalidation,valavi2022predictive}, we partition the globe into spatial block holdouts using S2 cells at levels 4 ($\approx 600$\,km), 6 ($\approx 150$\,km), 8 ($\approx 38$\,km), and 10 ($\approx 9$\,km). Cells are deterministically hashed into training (80\%), validation (10\%), and test (10\%) sets.

For model selection and hyperparameter tuning on held-out spatial blocks, standard metrics present key limitations: ROC-AUC saturates over continental extents because distant true negatives suppress false positive rates and does not punish oversmoothed maps \citep{lobo2008auc,jimenez2012insights}, while average precision is volatile for rare species with few positives \citep{boyd2013area,sofaer2019area}. In contrast, NDCG@$K$ truncates evaluation to the top $K$ cells with a logarithmic rank discount $1/\log_2(r+1)$, smoothing the ranking signal \citep{jarvelin2002cumulated,wang2013theoretical}. We select $K=500$, which maximizes cross-subset score correlation. To correct for prevalence-dependent baselines and avoid near-zero denominator instability on rare species ($\pi_s \to 0$), we compute the pooled ratio-of-sums \emph{skill score}:
\begin{equation}
    \text{Skill}_{\text{NDCG}@K} = \frac{\sum_s \big(\text{NDCG}_s@K - \mathbb{E}[\text{NDCG}_s@K]\big)}{\sum_s \big(1 - \mathbb{E}[\text{NDCG}_s@K]\big)}.
\end{equation}
For computational efficiency, NDCG@500 is evaluated on approximately half a million holdout locations across a random sample of 4{,}096 species, rather than the full holdout set. The block-level holdout ensures that all training data within the surrounding S2 cell has been removed, preventing spatial autocorrelation from inflating extrapolation scores.

While the training objective optimizes relative occurrence conditioned on location ($S \mid X$) to cancel unknown spatial sampling effort via target-group conditioning, our spatial holdout evaluation measures per-species geographic ranking ($X \mid S$). By Bayes' rule, the posterior spatial distribution satisfies $P(X \mid S) \propto P(S \mid X) \, P(X)$. Evaluating spatial NDCG directly on model outputs $\sigma(z_s(x))$ corresponds to imposing a uniform geographic prior ($P_{\text{eval}}(X) = c$) across candidate locations.

Our evaluation protocol enforces this prior in two ways: first, we omit cell-temperature sampling and evaluate uniformly over held-out S2 blocks; second, we binarize ground-truth targets ($y_{x,s} \in \{0, 1\}$). Binarization strips the heavy-tailed observation count confound from the evaluation metric, ensuring that presence in a remote, single-observation wilderness cell is rewarded equally to presence in a heavily visited observation hotspot, strictly testing the model's environmental discrimination.

For interpolation, we evaluate on three benchmarks from SINR \citep{cole2023spatial}: (1) the \textbf{geo-prior} task, where the SDM serves as a geographic prior for species classification; (2) eBird \textbf{S\&T range maps}, evaluated via mean average precision (MAP) over 535 North American bird species; and (3) \textbf{IUCN range maps}, evaluated via MAP over 2,418 species globally. Because S\&T range maps are derived from eBird data (which is part of GBIF), this benchmark evaluates within-distribution geographic fidelity rather than extrapolation to unseen regions. The eBird S\&T and IUCN benchmarks evaluate binary range prediction, a task OCELOT was not specifically trained for; we include them to show that a relative occurrence model can still produce effective geographic discrimination and outperform existing neural baselines on standard range benchmarks.

\subsection{Main Results and Ablations}
\label{sec:main_results}
\label{sec:ablations}

Table~\ref{tab:master_results} presents our primary experimental results and systematic ablation studies across all benchmarks. Entries report the mean $\pm$ standard deviation across $N=5$ independent random seeds at 5M training steps.

\begin{table*}[t]
\centering
\small
\setlength{\tabcolsep}{6pt}
\renewcommand{\arraystretch}{0.96}
\caption{\textbf{Benchmark and ablation results.} Spatial extrapolation evaluated via NDCG@500 Skill Score (averaged across S2 levels 4--10); interpolation evaluated via GeoPrior Top-1 (\%), eBird S\&T MAP (\%), and IUCN MAP (\%). Entries report mean $\pm$ standard deviation across $N=5$ seeds at 5M steps. Bold entries indicate top performers and those not statistically significantly different ($p \ge 0.05$, paired two-sided $t$-test across seeds).}
\label{tab:master_results}
\begin{threeparttable}
\begin{tabular}{lcccc}
\toprule
\textbf{Configuration} & \textbf{NDCG} & \textbf{GeoPrior (\%)} & \textbf{S\&T (\%)} & \textbf{IUCN (\%)} \\
\midrule
\multicolumn{5}{l}{\textit{Published Baselines}} \\
SINR (AN-full, spatial) & $0.036$\tnote{*} & 81.5 & 77.15 & 65.84 \\
SINR (AN-full) & --- & 81.9 & 80.48 & 76.07 \\
HSR-SDM (spatial) & --- & --- & 78.47 & 70.32 \\
\midrule
\multicolumn{5}{l}{\textit{Input Modalities}} \\
Spatial only (SH $L\!=\!50$) & $0.076 \pm 0.002$ & $86.2 \pm 0.0$ & $83.68 \pm 0.11$ & $75.09 \pm 0.08$ \\
BioClim only & $0.062 \pm 0.002$ & $85.3 \pm 0.0$ & $79.55 \pm 0.05$ & $74.98 \pm 0.07$ \\
Spatial + BioClim & $0.083 \pm 0.002$ & $86.3 \pm 0.0$ & $\mathbf{84.55 \pm 0.10}$ & $\mathbf{75.59 \pm 0.14}$ \\
AEF only (Satellite) & $0.090 \pm 0.002$ & $85.1 \pm 0.0$ & $77.27 \pm 0.11$ & $71.63 \pm 0.06$ \\
Spatial + AEF (OCELOT) & $\mathbf{0.115 \pm 0.002}$ & $\mathbf{86.4 \pm 0.0}$ & $83.22 \pm 0.07$ & $74.16 \pm 0.02$ \\
Spatial + AEF + BioClim & $\mathbf{0.115 \pm 0.001}$ & $\mathbf{86.5 \pm 0.0}$ & $83.23 \pm 0.06$ & $74.05 \pm 0.13$ \\
\midrule
\multicolumn{5}{l}{\textit{Taxonomic Hierarchy Readout}} \\
Species only ($K\!=\!1$) & $0.113 \pm 0.002$ & $86.4 \pm 0.0$ & $\mathbf{83.33 \pm 0.03}$ & $\mathbf{74.43 \pm 0.12}$ \\
Up to family ($K\!=\!3$, OCELOT) & $\mathbf{0.115 \pm 0.002}$ & $86.4 \pm 0.0$ & $83.22 \pm 0.07$ & $74.16 \pm 0.02$ \\
Up to class ($K\!=\!5$) & $\mathbf{0.115 \pm 0.001}$ & $\mathbf{86.5 \pm 0.0}$ & $83.19 \pm 0.03$ & $74.11 \pm 0.09$ \\
\midrule
\multicolumn{5}{l}{\textit{Spherical Harmonics Degree}} \\
Degree $L=30$ & $\mathbf{0.116 \pm 0.001}$ & $\mathbf{86.5 \pm 0.0}$ & $\mathbf{83.70 \pm 0.07}$ & $\mathbf{74.44 \pm 0.12}$ \\
Degree $L=50$ (OCELOT) & $\mathbf{0.115 \pm 0.002}$ & $\mathbf{86.4 \pm 0.0}$ & $83.22 \pm 0.07$ & $74.16 \pm 0.02$ \\
Degree $L=70$ & $0.114 \pm 0.001$ & $\mathbf{86.4 \pm 0.0}$ & $82.70 \pm 0.05$ & $73.47 \pm 0.05$ \\
\midrule
\multicolumn{5}{l}{\textit{Cell Temperature Sampling Exponent}} \\
$T_{\text{cell}} = 0.0$ (Uniform point) & $0.107 \pm 0.001$ & $\mathbf{86.6 \pm 0.0}$ & $80.77 \pm 0.10$ & $72.85 \pm 0.12$ \\
$T_{\text{cell}} = 0.50$ (OCELOT) & $\mathbf{0.115 \pm 0.002}$ & $86.4 \pm 0.0$ & $83.22 \pm 0.07$ & $\mathbf{74.16 \pm 0.02}$ \\
$T_{\text{cell}} = 1.00$ (Volumetric linear) & $0.107 \pm 0.001$ & $85.5 \pm 0.0$ & $\mathbf{84.21 \pm 0.07}$ & $72.78 \pm 0.15$ \\
\midrule
\multicolumn{5}{l}{\textit{Trunk Depth Scaling}} \\
Trunk depth 2 & $0.114 \pm 0.001$ & $86.4 \pm 0.0$ & $\mathbf{83.37 \pm 0.05}$ & $\mathbf{74.33 \pm 0.08}$ \\
Trunk depth 4 (OCELOT) & $0.115 \pm 0.002$ & $86.4 \pm 0.0$ & $83.22 \pm 0.07$ & $74.16 \pm 0.02$ \\
Trunk depth 8 & $\mathbf{0.116 \pm 0.001}$ & $\mathbf{86.5 \pm 0.0}$ & $82.97 \pm 0.07$ & $73.71 \pm 0.12$ \\
\bottomrule
\end{tabular}
\begin{tablenotes}[flushleft]
\footnotesize
\item[*] Since the SINR model does not support the same taxa, this model was evaluated on a random subset of 4\,096 species supported by both models. OCELOT achieves 0.1034 on the same evaluation set. No SINR checkpoints supporting environmental covariates were available.
\end{tablenotes}
\end{threeparttable}
\end{table*}

\paragraph{Feature modalities and environmental inputs.}
As shown in Table~\ref{tab:master_results}, our spatial-only model already outperforms all published coordinate-only baselines (SINR, \citealt{cole2023spatial}; HSR-SDM, \citealt{yuan2024hybrid}) across all benchmarks and exceeds even the environmental SINR baseline on GeoPrior and eBird S\&T, reflecting the combined effect of our training dataset, data pipeline, architecture, and loss formulation. However, coordinates alone cannot extrapolate across spatial holdouts: adding AlphaEarth Foundations (AEF) satellite embeddings improves spatial holdout ranking by 51\% over coordinates alone, with the largest relative gain (+68\%) emerging at the hardest 600\,km scale (Appendix~\ref{app:sh_scales}). Even without spatial coordinates, AEF embeddings alone outperform both BioClim-only and coordinate-only models ($0.090$ vs.\ $0.062$ and $0.076$ NDCG), as AEF representations implicitly encode geographic location, elevation, and climatic regimes \citep{brown2025alphaearth}. Furthermore, adding BioClim alongside AEF provides no measurable improvement, confirming that foundation embeddings render hand-crafted climatic variables redundant. Finally, this resolves the apparent paradox in prior studies \citep{cole2023spatial}, which concluded environmental covariates were unhelpful: standard range-map benchmarks evaluate coarse, macro-scale boundaries---where eBird S\&T explicitly incorporates bioclimatic predictors \citep{fink2020modeling} and IUCN polygons delineate broad extent-of-occurrence envelopes that penalize fine-scale habitat discrimination (explaining why coarse BioClim features slightly outscore fine-grained AEF on IUCN polygons). On these coarse targets, satellite features show little apparent gain, whereas true spatial extrapolation reveals their critical importance.

\paragraph{Taxonomic hierarchy readout.}
Decomposing species weights into hierarchical taxonomic embeddings consistently improves spatial extrapolation over flat species-only readouts ($K=1$), particularly across data-scarce rare species ($0.072 \to 0.075$, paired $p = 0.015$; Appendix~\ref{app:taxonomy_abundance}). Deeper clades ($K \ge 3$) show diminishing returns. We adopt $K=3$ (Family) as our default, which aligns with phylogenetic niche conservatism at the family rank \citep{crisp2009phylogenetic,wiens2005niche}.

\paragraph{Sampling temperature and architecture scaling.}
Varying cell temperature confirms $T_{\text{cell}} = 0.5$ as the empirical optimum ($0.115$ NDCG), outperforming both uniform point sampling ($T=0$, $0.107$) and linear volumetric sampling ($T=1$, $0.107$) by balancing between over-weighting noisy singletons and over-concentrating gradients on urban observation hotspots (Appendix~\ref{app:cell_temperature}). Trunk depth exhibits minimal sensitivity, with 8 layers offering only a marginal gain over 2 layers ($0.116$ vs.\ $0.114$ NDCG). When sweeping harmonic degree $L \in \{30, \dots, 70\}$, lower degrees ($L=30$) favor continental-scale extrapolation across expansive data voids, whereas higher degrees ($L \ge 50$) improve fine-grained ranking at sensor-relevant landscape scales ($10\text{--}40$\,km; Appendix~\ref{app:sh_scales}). In effect, the spherical harmonics degree determines the effective spatial resolution ($\sim C/2L$\,km): higher degrees resolve fine geographic boundaries but risk unconstrained oscillations across unobserved regions. We select $L=50$ as a balanced trade-off between local geographic detail and void stability.

\section{Conclusion and Future Work}
\label{sec:conclusion}

We presented a framework for scaling relative occurrence modeling to 100,000 species on 2 billion GBIF records by combining satellite foundation embeddings, spherical harmonic spatial features, an additive taxonomic readout, independent sigmoid losses with class-prior offsets, and a spatially grounded data pipeline. Satellite foundation embeddings proved essential for geographic extrapolation---substantially improving spatial holdout performance over coordinate-only models and rendering classical bioclimatic rasters redundant---demonstrating that prior findings of coordinate sufficiency were an artifact of interpolation-only evaluations. Notably, AEF embeddings alone outperform both coordinate-only and BioClim-only models on spatial extrapolation without any geographic coordinates. An additive taxonomic readout with $K=3$ levels (species, genus, family) improves predictions for rare taxa by sharing representations through the phylogenetic hierarchy. The resulting model outperforms existing neural baselines on GeoPrior and eBird S\&T while remaining competitive on coarse IUCN range maps and establishing strong geographic extrapolation capability on spatial holdouts. Although many factors differ between our system and prior baselines---training data scale, architecture, and loss formulation---the results demonstrate that a model optimized for geographic extrapolation need not sacrifice interpolation performance.

Several limitations remain. The model does not account for temporal dynamics: seasonal migration, phenological shifts, and long-term range changes are not modeled. Additionally, whereas classical bioclimatic variables are directly interpretable as physical quantities, latent satellite embeddings require post-hoc probing to attribute predictions to specific environmental drivers. Finally, the current system outputs a single static model without a clear way of adding additional (expert) observations or adding new species without re-training. Future directions include integrating temporal phenology, evaluating the model on real-world deployment data, and extending the training and evaluation to additional taxa and geographic regions.

\subsubsection*{Reproducibility Statement}
All code for data ingestion, model architectures, loss formulations, and evaluation metrics is implemented in JAX. Hyperparameter configurations, optimization schedules, and random seeds are documented in Section~\ref{sec:experiments} and Appendix~\ref{app:cell_temperature}. We plan to release the model weights.

\subsubsection*{Ethics Statement}
High-resolution species distribution predictions carry dual-use risks, particularly the localization of endangered species vulnerable to poaching or illegal harvesting. GBIF applies location generalization to flagged sensitive taxa following international biodiversity protocols, and OCELOT inherits these protections.

\subsubsection*{AI Use Statement}
The authors acknowledge the use of large language models as writing assistants for coding, literature summarization, grammatical refinement, and LaTeX formatting. All conceptual formulations and manuscript conclusions were conceived, executed, and verified by the authors.

\subsubsection*{Acknowledgements}

We are grateful to Rishabh Baghel, Federico Barbero, Michelangelo Conserva and Peter Veličković for their contributions to earlier versions of this work.

\bibliography{iclr2026_conference}

@article{phillips2006maximum,
  title={Maximum entropy modeling of species geographic distributions},
  author={Phillips, Steven J and Anderson, Robert P and Schapire, Robert E},
  journal={Ecological Modelling},
  volume={190},
  number={3-4},
  pages={231--259},
  year={2006},
  publisher={Elsevier}
}

@article{elith2009species,
  title={Species distribution models: ecological explanation and prediction across space and time},
  author={Elith, Jane and Leathwick, John R},
  journal={Annual Review of Ecology, Evolution, and Systematics},
  volume={40},
  pages={677--697},
  year={2009},
  publisher={Annual Reviews}
}

@article{fick2017worldclim,
  title={WorldClim 2: new 1-km spatial resolution climate surfaces for global land areas},
  author={Fick, Stephen E and Hijmans, Robert J},
  journal={International Journal of Climatology},
  volume={37},
  number={12},
  pages={4302--4315},
  year={2017},
  publisher={Wiley Online Library}
}

@incollection{botella2018deep,
  title={A deep learning approach to species distribution modelling},
  author={Botella, Christophe and Joly, Alexis and Bonnet, Pierre and Monestiez, Pascal and Munoz, Fran{\c{c}}ois},
  booktitle={Multimedia tools and applications for environmental \& biodiversity informatics},
  pages={169--199},
  year={2018},
  publisher={Springer}
}

@article{phillips2009sample,
  title={Sample selection bias and presence-only distribution models: implications for background and pseudo-absence data},
  author={Phillips, Steven J and Dud{\'\i}k, Miroslav and Elith, Jane and Graham, Catherine H and Lehmann, Anthony and Leathwick, John and Ferrier, Simon},
  journal={Ecological Applications},
  volume={19},
  number={1},
  pages={181--197},
  year={2009},
  publisher={Wiley Online Library}
}

@article{warton2010poisson,
  title={Poisson point process models solve the ``pseudo-absence problem'' for presence-only data in ecology},
  author={Warton, David I and Shepherd, Leah C},
  journal={The Annals of Applied Statistics},
  volume={4},
  number={3},
  pages={1383--1402},
  year={2010},
  publisher={Institute of Mathematical Statistics}
}

@inproceedings{gutmann2010noise,
  title={Noise-contrastive estimation: A new estimation principle for unnormalized statistical models},
  author={Gutmann, Michael and Hyv{\"a}rinen, Aapo},
  booktitle={Proceedings of the Thirteenth International Conference on Artificial Intelligence and Statistics},
  pages={297--304},
  year={2010}
}

@article{fithian2015bias,
  title={Bias correction in species distribution models: pooling survey and collection data for multiple species},
  author={Fithian, William and Elith, Jane and Hastie, Trevor and Keith, David A},
  journal={Methods in Ecology and Evolution},
  volume={6},
  number={4},
  pages={424--438},
  year={2015},
  publisher={Wiley Online Library}
}

@article{renner2015point,
  title={Point process models for presence-only analysis},
  author={Renner, Ian W and Elith, Jane and Baddeley, Adrian and Fithian, William and Hastie, Trevor and Phillips, Steven J and Popovic, Gordana and Warton, David I},
  journal={Methods in Ecology and Evolution},
  volume={6},
  number={4},
  pages={366--379},
  year={2015},
  publisher={Wiley Online Library}
}

@article{barber2021target,
  title={Target-group backgrounds prove effective at correcting sampling bias in {Maxent} models},
  author={Barber, Robert A and Ball, Stuart G and Morris, Roger KA and Gilbert, Francis},
  journal={Diversity and Distributions},
  volume={28},
  number={1},
  pages={126--141},
  year={2021},
  publisher={Wiley Online Library}
}

@article{valavi2022predictive,
  title={Predictive performance of presence-only species distribution models: a benchmark},
  author={Valavi, Roozbeh and Elith, Jane and Lahoz-Monfort, Jos{\'e} J and Guillera-Arroita, Gurutzeta},
  journal={Ecological Monographs},
  volume={92},
  number={1},
  pages={e01486},
  year={2022},
  publisher={Wiley Online Library}
}

@inproceedings{cole2023spatial,
  title={Spatial implicit neural representations for global-scale species mapping},
  author={Cole, Elijah and Van Horn, Grant and Lange, Christian and Shepard, Alexander and Leary, Patrick and Perona, Pietro and Loarie, Scott and Mac Aodha, Oisin},
  booktitle={International conference on machine learning},
  pages={6320--6342},
  year={2023},
  organization={PMLR}
}

@inproceedings{macaodha2019presence,
  title={Presence-only geographical priors for fine-grained image classification},
  author={Mac Aodha, Oisin and Cole, Elijah and Perona, Pietro},
  booktitle={2019 IEEE/CVF international conference on computer vision (ICCV)},
  pages={9595--9605},
  year={2019},
  organization={IEEE}
}

@inproceedings{russwurm2024geographic,
  title={Geographic location encoding with spherical harmonics and sinusoidal representation networks},
  author={Ru{\ss}wurm, Marc and Klemmer, Konstantin and Rolf, Esther and Zbinden, Robin and Tuia, Devis},
  booktitle={International Conference on Learning Representations},
  volume={2024},
  pages={1746--1759},
  year={2024}
}

@article{fink2020modeling,
  title={Modeling avian full annual cycle distribution and population trends with citizen science data},
  author={Fink, Daniel and Auer, Tom and Johnston, Alison and Ruiz-Gutierrez, Viviana and Hochachka, Wesley M and Kelling, Steve},
  journal={Ecological Applications},
  volume={30},
  number={3},
  pages={e02056},
  year={2020}
}

@article{yuan2024hybrid,
  title={Hybrid Spatial Representations for Species Distribution Modeling},
  author={Yuan, Shiran and Zhao, Hao},
  journal={arXiv preprint arXiv:2410.10937},
  year={2024}
}

@article{zbinden2024selection,
  title={On the selection and effectiveness of pseudo-absences for species distribution modeling with deep learning},
  author={Zbinden, Robin and van Tiel, Nina and Kellenberger, Benjamin and Hughes, Lloyd and Tuia, Devis},
  journal={Ecological Informatics},
  volume={81},
  pages={102623},
  year={2024},
  publisher={Elsevier}
}

@inproceedings{wang2020makes,
  title={What makes training multi-modal classification networks hard?},
  author={Wang, Weiyao and Tran, Du and Feiszli, Matt},
  booktitle={Proceedings of the IEEE/CVF Conference on Computer Vision and Pattern Recognition},
  pages={12695--12704},
  year={2020}
}

@article{vehtari2024pareto,
  title={Pareto smoothed importance sampling},
  author={Vehtari, Aki and Simpson, Daniel and Gelman, Andrew and Yao, Yuling and Gabry, Jonah},
  journal={Journal of Machine Learning Research},
  volume={25},
  number={72},
  pages={1--58},
  year={2024}
}

@article{ionides2008truncated,
  title={Truncated importance sampling},
  author={Ionides, Edward L},
  journal={Journal of Computational and Graphical Statistics},
  volume={17},
  number={2},
  pages={295--311},
  year={2008},
  publisher={Taylor \& Francis}
}

@article{lobo2008auc,
  title={{AUC}: a misleading measure of the performance of predictive distribution models},
  author={Lobo, Jorge M and Jim{\'e}nez-Valverde, Alberto and Real, Raimundo},
  journal={Global Ecology and Biogeography},
  volume={17},
  number={2},
  pages={145--151},
  year={2008},
  publisher={Wiley Online Library}
}

@article{jimenez2012insights,
  title={Insights into the area under the {ROC} curve ({AUC}) as a discrimination measure in species distribution modelling},
  author={Jim{\'e}nez-Valverde, Alberto},
  journal={Global Ecology and Biogeography},
  volume={21},
  number={4},
  pages={498--507},
  year={2012},
  publisher={Wiley Online Library}
}

@article{sofaer2019area,
  title={The area under the precision--recall curve as a performance metric for rare binary events},
  author={Sofaer, Helen R and Hoeting, Jennifer A and Jarnevich, Catherine S},
  journal={Methods in Ecology and Evolution},
  volume={10},
  number={4},
  pages={565--577},
  year={2019},
  publisher={Wiley Online Library}
}

@inproceedings{boyd2013area,
  title={Area under the precision-recall curve: point estimates and confidence intervals},
  author={Boyd, Kendrick and Eng, Kevin H and Page, C David},
  booktitle={European Conference on Machine Learning and Principles and Practice of Knowledge Discovery in Databases (ECML PKDD)},
  pages={451--466},
  year={2013},
  organization={Springer}
}

@article{jarvelin2002cumulated,
  title={Cumulated gain-based evaluation of {IR} techniques},
  author={J{\"a}rvelin, Kalervo and Kek{\"a}l{\"a}inen, Jaana},
  journal={ACM Transactions on Information Systems (TOIS)},
  volume={20},
  number={4},
  pages={422--446},
  year={2002},
  publisher={ACM New York, NY, USA}
}

@inproceedings{wang2013theoretical,
  title={A theoretical analysis of {NDCG} ranking measures},
  author={Wang, Yining and Wang, Liwei and Li, Yuanzhi and He, Di and Chen, Wei and Liu, Tie-Yan},
  booktitle={Proceedings of the 26th Annual Conference on Learning Theory (COLT)},
  pages={25--54},
  year={2013}
}

@article{shimatani2001multivariate,
  title={Multivariate point processes and spatial variation of species diversity},
  author={Shimatani, Kenichiro},
  journal={Forest Ecology and Management},
  volume={142},
  number={1-3},
  pages={215--229},
  year={2001},
  publisher={Elsevier}
}

@article{aiellolammens2015spthin,
  title={sp{T}hin: an {R} package for spatial thinning of species occurrence records for use in ecological niche models},
  author={Aiello-Lammens, Matthew E and Boria, Robert A and Radosavljevic, Aleksandar and Vilela, Bruno and Anderson, Robert P},
  journal={Ecography},
  volume={38},
  number={5},
  pages={541--545},
  year={2015},
  publisher={Wiley Online Library}
}

@article{steen2021evaluation,
  title={An evaluation of stringent filtering to improve species distribution models from citizen science data},
  author={Steen, Valerie A and Elphick, Chris S and Tingley, Morgan W},
  journal={Diversity and Distributions},
  volume={25},
  number={12},
  pages={1857--1869},
  year={2019},
  publisher={Wiley Online Library}
}

@article{moudry2024optimising,
  title={Optimising occurrence data in species distribution models: sample size, positional uncertainty, and sampling bias matter},
  author={Moudr{\'{y}}, V{\'{i}}t{\v{e}}zslav and Bazzichetto, Manuele and Remelgado, Ruben and Devillers, Rodolphe and Lenoir, Jonathan and Mateo, Rub{\'e}n G and Lembrechts, Jonas J and Sillero, Neftal{\'{\i}} and Lecours, Vincent and Cord, Anna F and others},
  journal={Ecography},
  volume={2024},
  number={12},
  pages={e07294},
  year={2024},
  publisher={Wiley Online Library}
}

@article{brown2025alphaearth,
  title={Alpha{E}arth {F}oundations: An embedding field model for accurate and efficient global mapping from sparse label data},
  author={Brown, Christopher F and Kazmierski, Michal R and Pasquarella, Valerie J and Rucklidge, William J and Samsikova, Maria and Zhang, Chenying and Shelhamer, Evan and Lahera, Eric and Wiles, Oliver and Ilyushchenko, Sergii and others},
  journal={arXiv preprint arXiv:2507.22291},
  year={2025}
}

@article{fithian2013finite,
  title={Finite-sample equivalence in statistical models for presence-only data},
  author={Fithian, William and Hastie, Trevor},
  journal={The Annals of Applied Statistics},
  volume={7},
  number={4},
  pages={1917--1939},
  year={2013},
  publisher={Institute of Mathematical Statistics}
}

@inproceedings{tan2020equalization,
  title={Equalization loss for long-tailed object recognition},
  author={Tan, Jingru and Wang, Changbao and Li, Buyu and Li, Quanquan and Ouyang, Wanli and Yin, Changqing and Yan, Junjie},
  booktitle={Proceedings of the IEEE/CVF Conference on Computer Vision and Pattern Recognition},
  pages={11662--11671},
  year={2020}
}

@inproceedings{zhai2023sigmoid,
  title={Sigmoid loss for language image pre-training},
  author={Zhai, Xiaohua and Mustafa, Basil and Kolesnikov, Alexander and Beyer, Lucas},
  booktitle={Proceedings of the IEEE/CVF International Conference on Computer Vision},
  pages={11975--11985},
  year={2023}
}

@inproceedings{mnih2012fast,
  title={A fast and simple algorithm for training neural probabilistic language models},
  author={Mnih, Andriy and Teh, Yee Whye},
  booktitle={International Conference on Machine Learning},
  pages={1751--1758},
  year={2012}
}

@inproceedings{kang2020decoupling,
  title={Decoupling representation and classifier for long-tailed recognition},
  author={Kang, Bingyi and Xie, Saining and Rohrbach, Marcus and Yan, Zhicheng and Gordo, Albert and Feng, Jiashi and Kalantidis, Yannis},
  booktitle={International Conference on Learning Representations (ICLR)},
  year={2020}
}

@article{mackenzie2002estimating,
  title={Estimating site occupancy rates when detection probabilities are less than one},
  author={MacKenzie, Darryl I and Nichols, James D and Lachman, Gideon B and Droege, Sam and Royle, J Andrew and Langtimm, Catherine A},
  journal={Ecology},
  volume={83},
  number={8},
  pages={2248--2255},
  year={2002},
  publisher={Wiley Online Library}
}

@inproceedings{menon2021longtail,
  title={Long-tail learning via logit adjustment},
  author={Menon, Aditya Krishna and Jayasumana, Sadeep and Rawat, Ankit Singh and Jain, Himanshu and Veit, Andreas and Kumar, Sanjiv},
  booktitle={International Conference on Learning Representations (ICLR)},
  year={2021}
}

@article{roberts2017crossvalidation,
  title={Cross-validation strategies for data with temporal, spatial, hierarchical or phylogenetic structure},
  author={Roberts, David R and Bahn, Volker and Ciuti, Simone and Boyce, Mark S and Elith, Jane and Guillera-Arroita, Gurutzeta and Hauenstein, Severin and Lahoz-Monfort, Jos{\'e} J and Schr{\"o}der, Boris and Thuiller, Wilfried and others},
  journal={Ecography},
  volume={40},
  number={8},
  pages={913--929},
  year={2017},
  publisher={Wiley Online Library}
}

@article{hadly2009niche,
  author  = {Hadly, Elizabeth A. and Spaeth, Paula A. and Li, Cheng},
  title   = {Niche conservatism above the species level},
  journal = {Proceedings of the National Academy of Sciences},
  year    = {2009},
  volume  = {106},
  number  = {Supplement 2},
  pages   = {19707--19714},
  doi     = {10.1073/pnas.0901648106}
}

@article{crisp2009phylogenetic,
  author  = {Crisp, Michael D. and Arroyo, Mary T. Kalin and Cook, Lyn G. and Gandolfo, Maria A. and Jordan, Gregory J. and McGlone, Matt S. and Weston, Peter H. and Westoby, Mark and Wilf, Peter and Linder, H. Peter},
  title   = {Phylogenetic biome conservatism on a global scale},
  journal = {Nature},
  year    = {2009},
  volume  = {458},
  number  = {7239},
  pages   = {754--756},
  doi     = {10.1038/nature07764}
}

@article{wiens2005niche,
  author  = {Wiens, John J. and Graham, Catherine H.},
  title   = {Niche Conservatism: Integrating Evolution, Ecology, and Conservation Biology},
  journal = {Annual Review of Ecology, Evolution, and Systematics},
  year    = {2005},
  volume  = {36},
  pages   = {519--539},
  doi     = {10.1146/annurev.ecolsys.36.102803.095431}
}

@article{kissling2012towards,
  title={Towards novel approaches to modelling biotic interactions in multispecies assemblages at large spatial extents},
  author={Kissling, W Daniel and Dormann, Carsten F and Groeneveld, J{\"u}rgen and Hickler, Thomas and K{\"u}hn, Ingolf and McInerny, Greg J and Montoya, Jos{\'e} M and R{\"o}mermann, Christine and Schiffers, Katja and Schurr, Frank M and Singer, Alexander and Svenning, Jens-Christian and Zimmermann, Niklaus E and O'Hara, Robert B},
  journal={Journal of Biogeography},
  volume={39},
  number={12},
  pages={2163--2178},
  year={2012},
  publisher={Wiley Online Library}
}

@article{wilkinson2021defining,
  title={Defining and evaluating predictions of joint species distribution models},
  author={Wilkinson, David P and Golding, Nick and Guillera-Arroita, Gurutzeta and Tingley, Reid and McCarthy, Michael A},
  journal={Methods in Ecology and Evolution},
  volume={12},
  number={3},
  pages={394--404},
  year={2021},
  publisher={Wiley Online Library}
}

@article{pollock2014understanding,
  title={Understanding co-occurrence by modelling species simultaneously with a {Joint Species Distribution Model} ({JSDM})},
  author={Pollock, Laura J and Tingley, Reid and Morris, William K and Golding, Nick and O'Hara, Robert B and Parris, Kirsten M and Vesk, Peter A and McCarthy, Michael A},
  journal={Methods in Ecology and Evolution},
  volume={5},
  number={5},
  pages={397--406},
  year={2014},
  publisher={Wiley Online Library}
}

@article{warton2015so,
  title={So many variables: joint modeling in community ecology},
  author={Warton, David I and Blanchet, F Guillaume and O'Hara, Robert B and Ovaskainen, Otso and Taskinen, Sara and Walker, Steven C and Hui, Francis KC},
  journal={Trends in Ecology \& Evolution},
  volume={30},
  number={12},
  pages={766--779},
  year={2015},
  publisher={Elsevier}
}

@article{meyer2015global,
  title={Global priorities for an effective information basis of biodiversity distributions},
  author={Meyer, Carsten and Kreft, Holger and Guralnick, Robert and Jetz, Walter},
  journal={Nature Communications},
  volume={6},
  number={1},
  pages={8221},
  year={2015},
  publisher={Nature Publishing Group}
}

@article{hughes2021sampling,
  title={Sampling biases shape our view of the natural world},
  author={Hughes, Alice C and Orr, Michael C and Ma, Keping and Costello, Mark J and Waller, John and Provoost, Pieter and Yang, Qi and Zhu, Chaodong and Qiao, Huijie},
  journal={Ecography},
  volume={44},
  number={9},
  pages={1259--1269},
  year={2021},
  publisher={Wiley Online Library}
}

@article{lomolino2004conservation,
  title={Conservation biogeography},
  author={Lomolino, Mark V and others},
  journal={Frontiers of Biogeography: new directions in the geography of nature},
  volume={293},
  pages={293--296},
  year={2004}
}

@article{wieczorek2005localizing,
  title={Localized spectral analysis on the sphere},
  author={Wieczorek, Mark A and Simons, Frederik J},
  journal={Geophysical Journal International},
  volume={162},
  number={3},
  pages={655--675},
  year={2005},
  publisher={Blackwell Publishing Ltd, Oxford, UK}
}

@article{kramerschadt2013importance,
  title={The importance of correcting for sampling bias in {MaxEnt} species distribution models},
  author={Kramer-Schadt, Stephanie and Niedballa, J{\"u}rgen and Pilgrim, John D and Schr{\"o}der, Boris and Lindenborn, Jana and Reinfelder, Vanessa and Stillfried, Milena and Heckmann, Ilja and Scharf, Anne K and Augeri, David M and others},
  journal={Diversity and Distributions},
  volume={19},
  number={11},
  pages={1366--1379},
  year={2013},
  publisher={Wiley Online Library}
}

@article{shimodaira2000improving,
  title={Improving predictive inference under covariate shift by weighting the log-likelihood function},
  author={Shimodaira, Hidetoshi},
  journal={Journal of Statistical Planning and Inference},
  volume={90},
  number={2},
  pages={227--244},
  year={2000},
  publisher={Elsevier}
}

@article{lee2021deduplicating,
  title={Deduplicating training data makes language models better},
  author={Lee, Katherine and Ippolito, Daphne and Nystrom, Andrew and Zhang, Chiyuan and Eck, Douglas and Callison-Burch, Chris and Carlini, Nicholas},
  booktitle={Proceedings of the 60th Annual Meeting of the Association for Computational Linguistics (Volume 1: Long Papers)},
  pages={8424--8445},
  year={2022}
}
\bibliographystyle{iclr2026_conference}

\newpage
\appendix
\section{Qualitative Relative Occurrence Maps}
\label{app:qualitative_maps}

\begin{figure}[!htbp]
\centering
\begin{tabular}{@{}p{0.47\textwidth}@{\hspace{0.05\textwidth}}p{0.47\textwidth}@{}}
\begin{subfigure}[t]{\linewidth}
  \vspace{0pt}
  \centering
  \includegraphics[width=0.88\linewidth]{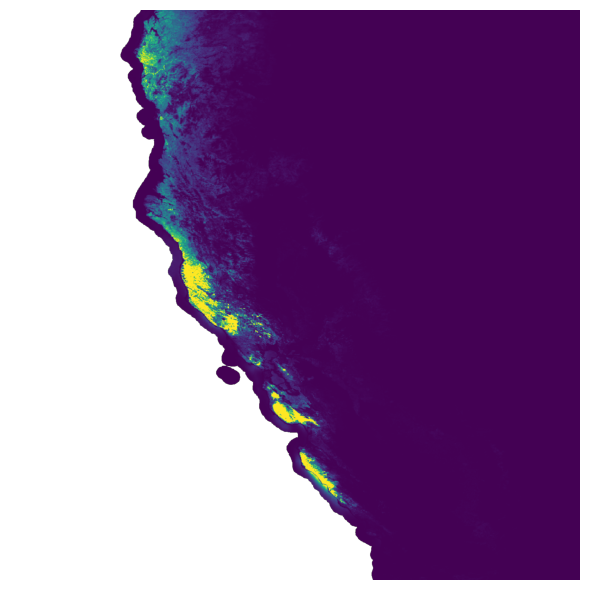}
  \caption{\textbf{Coast Redwood} (\emph{Sequoia sempervirens}). $9^\circ \times 9^\circ$ centered at $39.0^\circ\text{N}, 122.0^\circ\text{W}$ (Pacific Coast from southern Oregon to Point Conception, CA). Predicted relative occurrence probability is concentrated along coastal valleys and mountain slopes of the Pacific Coast Ranges, with near-zero probability across the interior Central Valley.}
  \label{fig:map_redwood}
\end{subfigure}
&
\begin{subfigure}[t]{\linewidth}
  \vspace{0pt}
  \centering
  \includegraphics[width=0.88\linewidth]{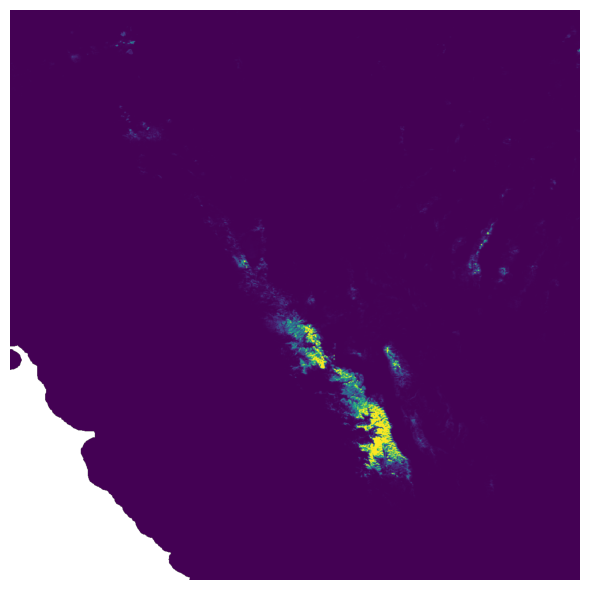}
  \caption{\textbf{American Pika} (\emph{Ochotona princeps}). $7^\circ \times 7^\circ$ centered at $38.5^\circ\text{N}, 119.5^\circ\text{W}$ (Sierra Nevada and western Great Basin). High values trace high-elevation terrain along the Sierra Nevada crest and the White Mountains, while the lower-elevation Central Valley and Owens Valley have near-zero relative occurrence probability.}
  \label{fig:map_pika}
\end{subfigure}
\\[0.5em]
\begin{subfigure}[t]{\linewidth}
  \vspace{0pt}
  \centering
  \includegraphics[width=0.88\linewidth]{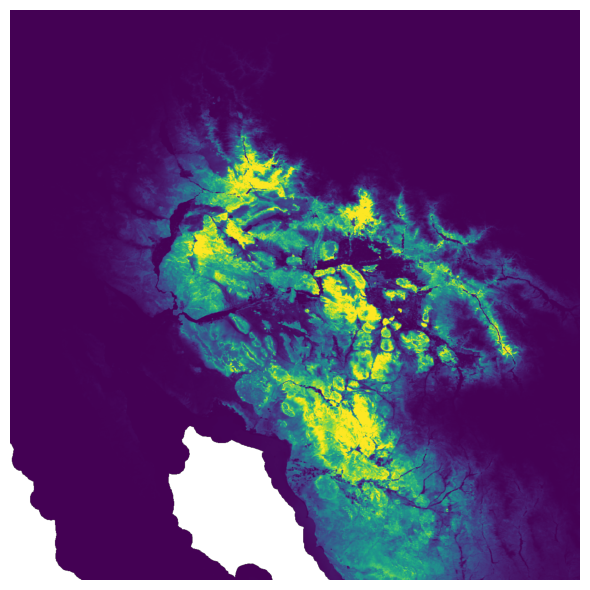}
  \caption{\textbf{Saguaro Cactus} (\emph{Carnegiea gigantea}). $7^\circ \times 7^\circ$ centered at $33.0^\circ\text{N}, 113.0^\circ\text{W}$ (Sonoran Desert in Arizona, Sonora, and lower Colorado River basin). Predicted relative occurrence probability covers low-elevation desert basins and bajadas, terminating northward at the higher-elevation Mogollon Rim.}
  \label{fig:map_saguaro}
\end{subfigure}
&
\begin{subfigure}[t]{\linewidth}
  \vspace{0pt}
  \centering
  \includegraphics[width=0.88\linewidth]{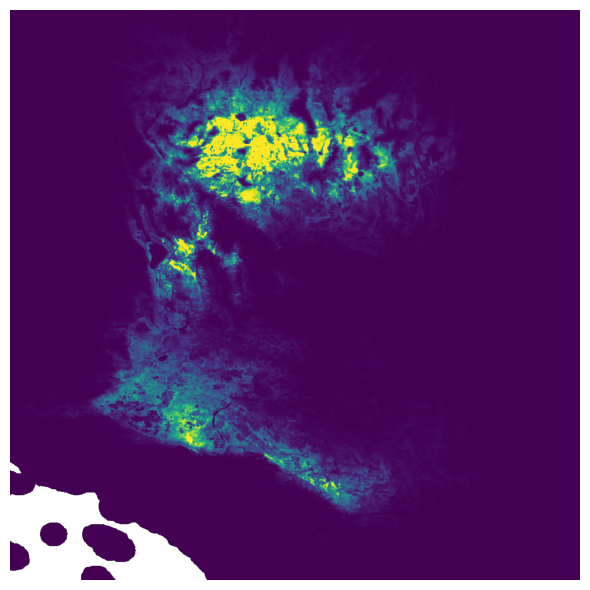}
  \caption{\textbf{Joshua Tree} (\emph{Yucca brevifolia}). $6^\circ \times 6^\circ$ centered at $36.0^\circ\text{N}, 116.5^\circ\text{W}$ (Mojave Desert and surrounding ranges). High relative occurrence probability covers the Mojave Desert plateau, bounded to the west by the Sierra Nevada and Tehachapi Mountains, with near-zero values in Death Valley and southern desert basins.}
  \label{fig:map_joshua_tree}
\end{subfigure}
\end{tabular}
\caption{Qualitative relative occurrence probability maps for four species evaluated across their geographic ranges. In each map, values represent the model's predicted relative occurrence probability. To resolve internal habitat contrast robustly across species with differing range sizes, $v_{\max}$ is set to the 95th percentile restricted to the top 95\% of the predicted probability mass. Warmer colors (yellow) indicate high relative occurrence probability, and dark blue denotes near-zero probability.}
\label{fig:appendix_species_maps}
\end{figure}
\clearpage

\section{Stratified Ablations: Taxonomic Depth and Harmonic Degree}
\label{app:stratified_ablations}

In this section, we provide detailed stratified breakdowns for two key architectural components: (1) the depth of the taxonomic hierarchy across species abundance terciles, and (2) the degree of the spherical harmonic spatial basis across varying spatial extrapolation scales.

\subsection{Taxonomic Hierarchy Readout Stratified by Species Abundance}
\label{app:taxonomy_abundance}

In Section~\ref{sec:ablations}, we observed that decomposing the final species readout into an additive sum of taxonomic embeddings ($\sum_{k=1}^K \mathbf{w}_{\tau_k(c)}$) improves overall spatial extrapolation performance over flat species-only readouts ($K=1$). However, aggregate evaluation across all 100,000 species conceals how the benefits of phylogenetic weight-sharing are distributed across common versus rare taxa.

To analyze this directly, we divide all evaluated species into equal abundance terciles based on their total observation frequency in the training corpus:
\begin{itemize}
    \item \textbf{Rare Tercile (Bottom 1/3):} Data-scarce species with the fewest observations in GBIF.
    \item \textbf{Intermediate Tercile (Middle 1/3):} Moderately sampled species.
    \item \textbf{Common Tercile (Top 1/3):} Highly sampled, ubiquitous species.
\end{itemize}

Table~\ref{tab:taxonomy_terciles} presents the spatial extrapolation NDCG@500 skill scores (averaged across all four spatial holdout levels: L4, L6, L8, and L10) across $N=5$ independent random seeds.

\begin{table}[h]
\centering
\small
\caption{\textbf{Taxonomic hierarchy depth sweep stratified by species abundance terciles.} Scores report mean NDCG@500 skill score $\pm$ standard deviation across $N=5$ independent seeds, averaged across S2 levels 4, 6, 8, and 10. Bold entries denote the top-performing configuration in each column and all configurations that are statistically indistinguishable from it ($p \ge 0.05$, paired two-sided $t$-test).}
\label{tab:taxonomy_terciles}
\vspace{0.5em}
\begin{tabular}{lcccc}
\toprule
\textbf{Hierarchy Depth ($K$)} & \textbf{Rare (Bottom 1/3)} & \textbf{Intermediate} & \textbf{Common (Top 1/3)} & \textbf{Overall Pooled} \\
\midrule
$K = 1$ (Species only) & $0.072 \pm 0.001$ & $0.080 \pm 0.000$ & $0.111 \pm 0.000$ & $0.094 \pm 0.000$ \\
$K = 2$ (Up to genus) & $0.074 \pm 0.001$ & $0.081 \pm 0.000$ & $\mathbf{0.112 \pm 0.001}$ & $0.094 \pm 0.000$ \\
$K = 3$ (Up to family, Ours) & $\mathbf{0.075 \pm 0.001}$ & $0.082 \pm 0.000$ & $\mathbf{0.112 \pm 0.000}$ & $\mathbf{0.095 \pm 0.001}$ \\
$K = 4$ (Up to order) & $\mathbf{0.075 \pm 0.001}$ & $\mathbf{0.083 \pm 0.000}$ & $\mathbf{0.112 \pm 0.000}$ & $0.095 \pm 0.000$ \\
$K = 5$ (Up to class) & $\mathbf{0.076 \pm 0.001}$ & $\mathbf{0.083 \pm 0.000}$ & $\mathbf{0.112 \pm 0.000}$ & $\mathbf{0.096 \pm 0.000}$ \\
\bottomrule
\end{tabular}
\end{table}

\paragraph{Key Findings:}
\begin{enumerate}
    \item \textbf{Rare taxa require taxonomic regularization:} Flat species embeddings ($K=1$) perform significantly worse than hierarchical models across all tiers ($p = 0.015$ vs.\ $K=3$ on rare taxa; $p < 0.001$ on intermediate taxa; $p = 0.002$ on common taxa). Hierarchical sharing ($K=1 \to K=3$) provides a consistent $+4.7\%$ gain ($0.072 \to 0.075$) across spatial holdout scales for rare taxa.
    \item \textbf{Diminishing returns beyond Genus and Family:} Once taxonomic representations are included ($K \ge 2$), performance differences across deeper clades ($K \in \{3, 4, 5\}$) are minimal ($\Delta \le 0.0006$ on rare taxa) and statistically tied with our default $K=3$ ($p = 0.793$ for $K=3$ vs.\ $K=4$; $p = 0.287$ for $K=3$ vs.\ $K=5$). 
    \item \textbf{Ecological justification for $K=3$:} In ecological theory, morphological and physiological tolerances are strongly conserved at the Family rank (e.g., \textit{Pinaceae} for boreal conifers, \textit{Cactaceae} for arid succulents, \textit{Formicidae} for ants) \citep{crisp2009phylogenetic,wiens2005niche}. We adopt $K=3$ (Species + Genus + Family) as our default because it captures the full statistical regularization benefit for rare species while maintaining biologically coherent clades.
\end{enumerate}

\subsection{Spherical Harmonics Degree Across Spatial Extrapolation Scales}
\label{app:sh_scales}

The spatial coordinate tower maps spherical coordinates $(\theta, \phi)$ to harmonic basis features $Y_\ell^m(\theta, \phi)$ up to maximum degree $L$, producing $(L+1)^2$ basis dimensions. On Earth's sphere (circumference $C \approx 40{,}075$\,km), the minimum spatial half-wavelength (the physical resolving limit of the basis) is given by \citep{wieczorek2005localizing}:
\begin{equation}
    d_{\text{res}}(L) = \frac{\pi R}{L} = \frac{C}{2L} \approx \frac{20{,}037\text{ km}}{L}.
\end{equation}

Table~\ref{tab:sh_scale_breakdown} breaks down the performance of spherical harmonic degree sweeps ($L \in \{30, \dots, 70\}$) across our four nested spatial evaluation scales: Level 4 (600\,km blocks), Level 6 (150\,km blocks), Level 8 (40\,km blocks), and Level 10 (10\,km blocks).

\begin{table}[h]
\centering
\small
\caption{\textbf{Spherical harmonics degree ablation broken down across spatial holdout scales.} Scores report mean NDCG@500 skill score $\pm$ standard deviation across $N=5$ independent seeds at 5M steps. Bold entries denote the top performer in each column and all configurations statistically indistinguishable from it ($p \ge 0.05$, paired two-sided $t$-test).}
\label{tab:sh_scale_breakdown}
\vspace{0.5em}
\begin{tabular}{lccc}
\toprule
\textbf{Degree ($L$)} & \textbf{Basis Resolution ($d_{\text{res}}$)} & \textbf{Level 4 (600\,km)} & \textbf{Level 6 (150\,km)} \\
\midrule
$L = 30$ & $\approx 668$\,km & $\mathbf{0.063 \pm 0.004}$ & $\mathbf{0.101 \pm 0.002}$ \\
$L = 40$ & $\approx 501$\,km & $\mathbf{0.060 \pm 0.003}$ & $\mathbf{0.100 \pm 0.002}$ \\
$L = 50$ (Ours) & $\approx 401$\,km & $0.057 \pm 0.002$ & $\mathbf{0.101 \pm 0.003}$ \\
$L = 60$ & $\approx 334$\,km & $0.053 \pm 0.004$ & $\mathbf{0.100 \pm 0.003}$ \\
$L = 70$ & $\approx 286$\,km & $0.050 \pm 0.003$ & $0.100 \pm 0.002$ \\
\midrule
\textbf{Degree ($L$)} & \textbf{Level 8 (40\,km)} & \textbf{Level 10 (10\,km)} & \textbf{Mean (Pooled)} \\
\midrule
$L = 30$ & $0.136 \pm 0.004$ & $0.163 \pm 0.002$ & $\mathbf{0.116 \pm 0.001}$ \\
$L = 40$ & $\mathbf{0.137 \pm 0.004}$ & $\mathbf{0.164 \pm 0.003}$ & $\mathbf{0.116 \pm 0.001}$ \\
$L = 50$ (Ours) & $\mathbf{0.137 \pm 0.003}$ & $0.164 \pm 0.003$ & $\mathbf{0.115 \pm 0.002}$ \\
$L = 60$ & $\mathbf{0.137 \pm 0.003}$ & $\mathbf{0.165 \pm 0.002}$ & $0.114 \pm 0.001$ \\
$L = 70$ & $\mathbf{0.138 \pm 0.003}$ & $\mathbf{0.166 \pm 0.003}$ & $0.114 \pm 0.001$ \\
\bottomrule
\end{tabular}
\end{table}

\paragraph{Trade-off Between Continental Extrapolation and Field Utility:}
\begin{enumerate}
    \item \textbf{Macro-scale extrapolation ($L=30$):} At continental scales (Level 4, 600\,km blocks), lower degrees perform best ($0.063$ for $L=30$ vs.\ $0.050$ for $L=70$). In global biodiversity data, persistent geographic sampling voids (the \textit{Wallacean shortfall}, such as the central Congo Basin and Amazonian interfluves) have characteristic diameters of $500\text{--}1{,}000$\,km \citep{meyer2015global,hughes2021sampling,lomolino2004conservation}. Setting $L=30$ ($d_{\text{res}} \approx 668$\,km) provides an exact spatial Nyquist match to these void dimensions, preventing unconstrained high-frequency oscillations across large data gaps.
    \item \textbf{Field deployment utility ($L=50$):} In practical conservation applications, autonomous sensors (camera traps, passive acoustic monitors, eDNA transects) are deployed within reserves, corridors, and protected areas at landscape scales ($10\text{--}40$\,km, corresponding to S2 Levels 8 and 10). At these operational scales, higher harmonic bandwidth monotonically improves spatial ranking ($0.163 \to 0.166$ on Level 10) by resolving topography, elevational gradients, and reserve perimeters.
    \item \textbf{Conclusion:} Setting $L=50$ ($d_{\text{res}} \approx 401$\,km) provides a balance: it captures over 95\% of the fine-scale resolution benefits ($0.164$ on Level 10) without suffering from the continental void overfitting of $L=70$.
\end{enumerate}

\section{Data Resampling}
\label{app:cell_temperature}

\subsection{Pareto Heavy Tail and the Necessity of Resampling}
\label{app:subsec_pareto_tail}

Citizen-science observation volume exhibits heavy geographic clustering. Across our training dataset of $198{,}094{,}492$ aggregated spatial cells, cell observation counts $N_i$ span from $1$ to $557{,}649$ (median $N = 1$, mean $N = 10.48$).

We model the upper tail of cell observation counts as a Pareto distribution:
\begin{equation}
P(N > x) = \left( \frac{x_{\min}}{x} \right)^\alpha, \quad x \ge x_{\min},
\label{eq:pareto_tail}
\end{equation}
where $\alpha > 0$ is the tail index. We estimate $\alpha$ using the maximum likelihood Hill estimator:
\begin{equation}
\hat{\alpha}_{\text{Hill}} = \left( \frac{1}{K} \sum_{k=1}^K \ln \frac{N_{(n-k+1)}}{x_{\text{threshold}}} \right)^{-1},
\label{eq:hill_estimator}
\end{equation}
where $N_{(1)} \le \dots \le N_{(n)}$ are the order statistics and $K$ is the number of cells in the tail. Table~\ref{tab:hill_tail_analysis} presents the estimated tail index $\hat{\alpha}$ across upper percentiles of our training dataset. Across the upper tail, $\hat{\alpha}$ remains concentrated between $0.89$ and $0.97$ (with minimum Kolmogorov-Smirnov distance identifying $N \ge 60$ as the optimal power-law cutoff, yielding $\hat{\alpha} = 0.8940$).

This empirical analysis shows that the training data is Pareto heavy-tailed with $\alpha < 2$. In standard stochastic gradient descent with loss weighting (where cells are sampled uniformly and their loss terms are scaled by observation volume $N_i$), the variance of the mini-batch gradient estimator scales with the second moment of the weights: $\mathbb{E}[\|\nabla \mathcal{L}\|^2] \propto \mathbb{E}[N_i^2]$. For a Pareto distribution, the second moment converges if and only if $\alpha > 2$. Because $\alpha < 2$, the second moment diverges ($\mathbb{E}[N_i^2] = \infty$), resulting in unbounded gradient variance driven by rare, hyper-frequented hotspot cells \citep{ionides2008truncated,vehtari2024pareto}. Weighted resampling resolves this instability by incorporating cell observation frequencies directly into the proposal distribution ($w_i \propto N_i^{T_{\text{cell}}}$) rather than scaling loss gradients.

\begin{table}[h]
\centering
\small
\caption{Hill estimator of the Pareto tail exponent $\alpha$ across upper percentiles of the $198{,}094{,}492$ training cell observation counts $N$.}
\label{tab:hill_tail_analysis}
\begin{tabular}{lccc}
\toprule
\textbf{Tail Cutoff} & \textbf{Threshold $N$} & \textbf{Cells in Tail ($K$)} & \textbf{Tail Index $\hat{\alpha}$} \\
\midrule
Top 5.0\% (95th) & $N \ge 18$ & $10{,}350{,}675$ & $0.9411$ \\
Top 2.0\% (98th) & $N \ge 49$ & $3{,}984{,}139$ & $0.8968$ \\
Top 1.0\% (99th) & $N \ge 109$ & $1{,}981{,}538$ & $0.9110$ \\
Top 0.5\% (99.5th) & $N \ge 245$ & $991{,}515$ & $0.9724$ \\
\bottomrule
\end{tabular}
\end{table}

\subsection{Cell Weighting}
\label{app:subsec_spatial_extrapolation}

\begin{figure}[h]
\centering
\includegraphics[width=0.60\linewidth]{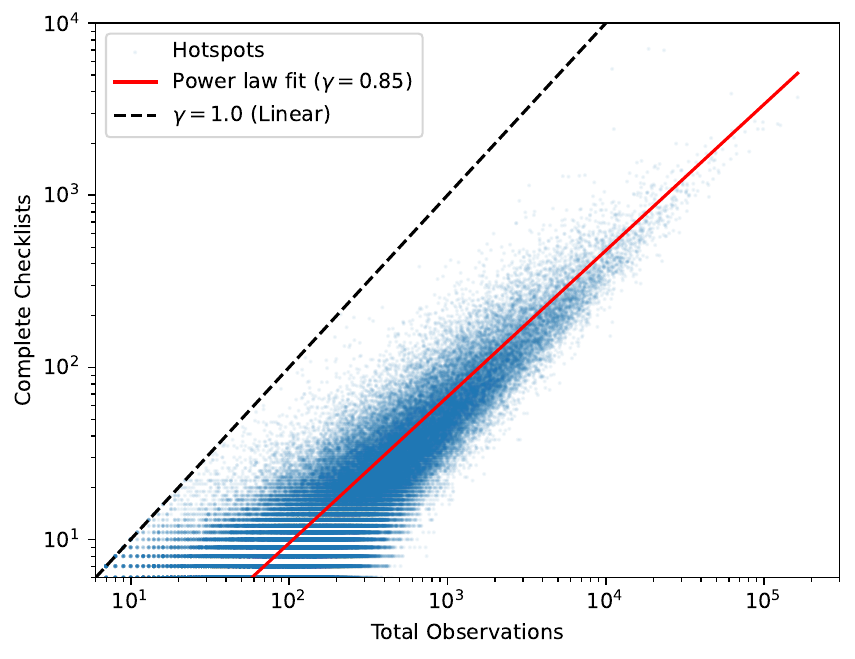}
\caption{\textbf{Empirical relationship between observation counts and observer effort.} Evaluated across 122,548 eBird hotspots. Complete checklists submitted scale as a power law ($B \propto N^{0.85}$, $R^2 = 0.987$) relative to total sightings. Setting $T_{\text{cell}} = 0.50$ shifts the training objective toward the uniform deployment prior.}
\label{fig:ebird_powerlaw}
\end{figure}

In Section~\ref{sec:ablations}, hyperparameter optimization identified $T_{\text{cell}} \approx 0.50$ as the empirical optimum for spatial extrapolation, bringing the training distribution closer to a uniform spatial prior. Beyond covarying observer effort, two complementary mechanisms justify sub-linear weighting:
\begin{enumerate}
    \item \textbf{Biodiversity hotspot correction:} A single visit to a biodiversity hotspot yields more presence records than a visit to a sparse habitat, conflating observer effort with local biodiversity and over-representing species-rich areas. Structured databases such as eBird provide both raw observation counts and independent checklist counts at the same locations, enabling direct calibration of this relationship: Figure~\ref{fig:ebird_powerlaw} demonstrates that across $122{,}548$ modeled eBird hotspots, physical sampling events (checklists) scale sub-linearly with raw observations ($\text{Effort} \propto N^{0.85}$, $R^2 = 0.987$). Sub-linear weighting helps correct for this over-representation.
    \item \textbf{Soft spatial deduplication:} Repeated observations from the same static 10\,m pixel provide diminishing marginal environmental information; sub-linear weighting acts as soft deduplication \citep{lee2021deduplicating}, downweighting redundant sightings without discarding rare species.
\end{enumerate}

\section{AlphaEarth Foundations: Spatial Resolution, Receptive Field, and Autocorrelation}
\label{app:aef_spatial_resolution}

In Section~\ref{sec:experimental_setup}, our model ingests 64-dimensional satellite embeddings from AlphaEarth Foundations (AEF; \citealt{brown2025alphaearth}) sampled on a global 10\,m spatial grid. In quantitative spatial ecology, a well-established principle dictates that the grain size $R$ of environmental predictor rasters should be at least twice the positional uncertainty $\epsilon$ of the biological occurrence records ($R \ge 2\epsilon$; \citealt{moudry2024optimising}) to avoid severe point-to-pixel misalignment and spurious fine-scale overfitting. Because opportunistic citizen-science observations (such as GBIF and iNaturalist records) frequently exhibit positional uncertainty ranging from tens of meters (GPS receiver noise) to several hundred meters (records snapped to trailheads, road access points, or administrative centroids), a direct 10\,m raster extraction might at first glance appear incongruent with ecological best practices.

In this section, we clarify our hypothesis that AEF embeddings do not suffer from point-to-pixel misalignment because they are inherently multi-scale representations.

\subsection{Backbone Architecture and Effective Receptive Field (\texorpdfstring{$1.28\text{ km}$}{1.28 km})}
\label{app:subsec_aef_receptive_field}

Unlike conventional environmental rasters (such as raw multispectral bands or vegetation indices), AEF embeddings are not localized point reflectance measurements. As documented by \citet{brown2025alphaearth}, the AEF embedding field model is trained on dense temporal sequences of multi-sensor satellite imagery (Sentinel-2 multispectral, Sentinel-1 synthetic aperture radar, and Landsat 8/9):
\begin{enumerate}
    \item \textbf{Input Tile Geometry:} Imagery is ingested as $128 \times 128$ pixel spatial patches at 10\,m ground sampling distance, covering a physical ground footprint of $1.28\text{ km} \times 1.28\text{ km}$ on Earth.
    \item \textbf{Global Spatial Self-Attention:} The encoder backbone comprises 15 repeated Space-Time-Precision (STP) blocks. Within each STP block, spatial information is processed via a downsampled tokenization operator equipped with full pairwise spatial self-attention, interleaved with multi-scale Laplacian pyramid cross-scale exchanges and local $3 \times 3$ convolutional operators \citep{brown2025alphaearth}.
    \item \textbf{Contextual Integration:} Because the spatial self-attention mechanism computes dot-product attention across all spatial tokens within the input tile, every individual 10\,m output embedding vector $\vx_{\text{env}} \in \mathbb{R}^{64}$ attends to and integrates multi-spectral, radar, and structural patterns across the \emph{entire} $1.28\text{ km} \times 1.28\text{ km}$ footprint.
\end{enumerate}
Consequently, while AEF embeddings are queryable at 10\,m grid spacing, each embedding represents an integrated landscape representation spanning $>1$\,km$^2$, operating well above the typical scale of observer GPS error.

\subsection{Empirical Semivariogram and Spatial Continuity}
\label{app:subsec_aef_semivariogram}

To empirically quantify the continuity of AEF embeddings across space, we evaluate their empirical semivariogram over a global distribution of terrestrial habitats.

\paragraph{Sampling Protocol.}
We sample 4,096 seed locations uniformly distributed across the globe using a spherical Fibonacci lattice. Filtering for seeds whose centroids intersect continental or insular land boundaries yields $1{,}502$ land disks. For each seed location $\mathbf{s}_k$, we construct a local disk of $N=128$ points spanning an angular radius of 10° ($\approx 1{,}110$\,km, diameter $\approx 2{,}220$\,km). We compute all pairwise great-circle geographic distances $d(\mathbf{x}_i, \mathbf{x}_j)$ and Euclidean embedding distances $\|\mathbf{e}(\mathbf{x}_i) - \mathbf{e}(\mathbf{x}_j)\|_2$, yielding $11{,}499{,}128$ valid pairwise comparisons across the globe.

\begin{figure}[h]
\centering
\includegraphics[width=0.65\linewidth]{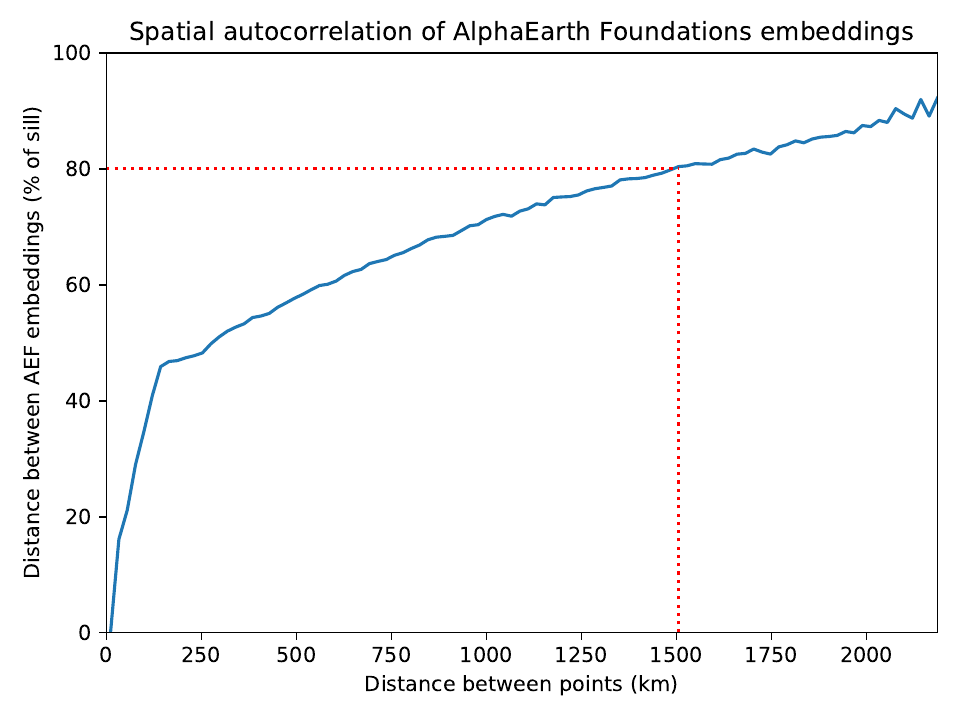}
\caption{\textbf{Empirical semivariogram of AlphaEarth Foundations embeddings.} Computed across $11{,}499{,}128$ pairwise point comparisons sampled from $1{,}502$ land-centroid disks globally (128 sample points per 10° disk) at native 10\,m resolution. The $y$-axis is scaled as a percentage of the theoretical sill ($1.4114$), representing the expected distance between independent uniform unit vectors on the hypersphere $\mathbb{S}^{63}$. The red dotted line marks the effective autocorrelation range of $1{,}505.49$\,km at 80\% of the sill ($1{,}175.82$\,km at 75\%).}
\label{fig:aef_semivariogram}
\end{figure}

\paragraph{Theoretical Sill on \texorpdfstring{$\mathbb{S}^{63}$}{S^63}.}
For two independent, uniformly distributed random unit vectors $\mathbf{u}, \mathbf{v} \in \mathbb{S}^{n}$ on the $n$-dimensional unit hypersphere (with $n = 64 - 1 = 63$), the expected Euclidean distance (the theoretical sill) can be solved analytically:
\begin{equation}
\text{Sill} = \mathbb{E}_{\mathbf{u}, \mathbf{v} \sim \text{Uniform}(\mathbb{S}^n)}\big[ \|\mathbf{u} - \mathbf{v}\|_2 \big] = \frac{2^n \, \Gamma\left(\frac{n+1}{2}\right)^2}{\sqrt{\pi} \, \Gamma\left(n + \frac{1}{2}\right)} \approx 1.4114.
\label{eq:hypersphere_sill}
\end{equation}
In Figure~\ref{fig:aef_semivariogram}, we rescale the vertical axis as a percentage of this theoretical sill, where 100\% represents complete spatial independence.

\paragraph{Empirical Findings.}
As shown in Figure~\ref{fig:aef_semivariogram}, AEF embeddings maintain spatial continuity across continental distances rather than decaying rapidly. This demonstrates that they are not merely hyper-local pixel measurements, but inherently multi-scale representations that preserve global and macro-ecological information.

\subsection{Implications for Species Distribution Modeling}
\label{app:subsec_aef_sdm_implications}

These structural and empirical properties have two key implications for our modeling framework:
\begin{enumerate}
    \item \textbf{Immunity to Centroid Snapping:} Because $\mathbf{e}(\mathbf{x})$ varies continuously across local topography and vegetation gradients, records that have been snapped to trailheads, county centroids, or public access points retain the representative environmental signature of the surrounding biome ($\mathbf{e}(\mathbf{x}_{\text{recorded}}) \approx \mathbf{e}(\mathbf{x}_{\text{true}})$).
    \item \textbf{Observer Noise Cancellation in SGD:} Across the 2.05 billion occurrences in our training corpus, local observer displacement vectors $\boldsymbol{\epsilon}$ are presumably approximately zero-mean. Because the loss is smooth with respect to $\mathbf{e}(\mathbf{x})$, stochastic gradient updates average out local positional errors in expectation over hundreds of thousands of sightings, preventing parameter collapse onto localized pixel artifacts.
\end{enumerate}

\section{Equivalence to Target-Group Backgrounds and the NCE Discriminant}
\label{app:tgb_nce}

This appendix establishes the equivalence between our multi-label sigmoid training and target-group background (TGB) pseudo-absence sampling \citep{phillips2009sample}, derives the asymptotic discriminant of each species' classifier via noise-contrastive estimation \citep{gutmann2010noise}, and identifies the residual distortion when species differ in their sampling effort.

\subsection{Setup}

Let $x \in \mathcal{D}$ denote a location, $\alpha_s(x) \geq 0$ the occurrence intensity (habitat suitability) of species $s$, and $\sigma_s(x) \geq 0$ the observer effort for species $s$ at $x$. The observation intensity is:
\begin{equation}
    n_s(x) = \sigma_s(x) \, \alpha_s(x).
    \label{eq:app_obs_intensity}
\end{equation}
Presence records for species $s$ are drawn with density proportional to $n_s(x)$. The SDM's goal is to recover $\alpha_s$ from these records, but observations are confounded by the unknown effort $\sigma_s$.

\subsection{Multi-label sigmoid BCE is TGB}

In our training procedure, each observed location $x$ where species $s$ was recorded contributes a positive example to classifier $s$ and a negative example to every other classifier $s' \neq s$. Unfolding across all training locations, the binary classifier for a focal species $y$ receives:
\begin{itemize}
    \item \textbf{Positives:} all locations where $y$ was observed, drawn from $n_y(x)$.
    \item \textbf{Negatives:} all locations where any species $z \neq y$ was observed, drawn from $\sum_{z \neq y} n_z(x)$.
\end{itemize}
This is identically the training data of a TGB classifier for species $y$ \citep{phillips2009sample}. Our multi-label formulation and TGB are not merely analogous---they partition the same global set of positive-negative contrastive pairs (every pair of observations belonging to different species) into per-species binary classification problems, and the two partitions produce the same training set for each species' classifier. The only difference is organizational: TGB groups by species, training one classifier at a time against pooled backgrounds; our formulation groups by location, updating all $S$ classifiers simultaneously at each observed site.

\subsection{The NCE discriminant}

By the convergence theorem for noise-contrastive estimation \citep{gutmann2010noise}, the Bayes-optimal discriminant of a binary classifier trained to distinguish a signal density from a noise density converges to their log density ratio. For species $y$, the signal density is $n_y(x)$ and the noise density is $\sum_{z \neq y} n_z(x)$. In the limit of sufficient model capacity and data:
\begin{equation}
    f_y(x) \;=\; \log n_y(x) \;-\; \log \sum_{z \neq y} n_z(x) \;+\; \text{const}.
    \label{eq:app_nce_discriminant}
\end{equation}

\paragraph{Under proportional bias.}
When all species share the same effort surface---$\sigma_s(x) = e(x)$ for all $s$, the assumption underlying TGB---the effort cancels from the density ratio:
\begin{equation}
    f_y(x) \;=\; \log \alpha_y(x) \;-\; \log \sum_{z \neq y} \alpha_z(x) \;+\; \text{const}.
    \label{eq:app_proportional_bias}
\end{equation}
The first term is the species' occurrence intensity in log space. The second term, $\log \sum_{z \neq y} \alpha_z(x)$, is the community normalization: it ensures that the outputs across species are interpretable as relative occurrence rates $P(Y = s \mid X = x, \text{observation at } x) = \alpha_s(x) / \sum_{s'} \alpha_{s'}(x)$, matching the estimand defined in Equation~\ref{eq:relative_occurrence} of the main text. In a single-species occupancy model---where the goal is to recover $\alpha_y(x)$ in isolation---this normalization is undesirable, as it suppresses predictions in species-rich areas. In our classification setting, where the target is relative occurrence, it is the correct normalizer.

\paragraph{When proportional bias fails.}
In practice, species differ in where they are preferentially detected: birders spend extra time in warbler habitat, herpetologists flip logs where reptiles shelter. Write the species-specific effort as $\sigma_s(x) = r_s(x) \bar{\sigma}(x)$, where $\bar{\sigma}(x)$ is the shared (species-independent) component and $r_s(x)$ is the species-specific effort ratio. Expanding Equation~\ref{eq:app_nce_discriminant}:
\begin{equation}
    f_y(x) \;=\; \log \alpha_y(x) \;-\; \log \sum_{z \neq y} \alpha_z(x) \;+\; \underbrace{\log \sigma_y(x) \;-\; \log \tilde{\sigma}(x)}_{\text{effort mismatch}} \;+\; \text{const},
    \label{eq:app_effort_mismatch}
\end{equation}
where $\tilde{\sigma}(x) = \frac{\sum_{z \neq y} \sigma_z(x) \alpha_z(x)}{\sum_{z \neq y} \alpha_z(x)}$ is the suitability-weighted pool effort. The effort mismatch term captures the difference between the focal species' effort and the community-averaged effort. It vanishes when proportional bias holds ($r_s = \text{const}$ for all $s$) and is the irreducible cost of presence-only data: removing it would require detection/non-detection records that identify species-specific effort directly \citep{mackenzie2002estimating}.

\subsection{Spatial Ranking and Invariance of Binary Habitat Metrics}
\label{app:ranking_invariance}

A natural question is whether training on relative occurrence $P(Y = y \mid X = x) \propto \alpha_y(x) / R(x)$, where $R(x) = \sum_s \alpha_s(x)$ is total community richness, distorts the spatial ranking of locations for a given species. 

Because $R(x)$ varies across biomes, local variations in community richness can indeed reorder locations within a species' habitat: a moderate-suitability site in a species-poor region can yield a higher relative occurrence score than a high-suitability site in a megadiverse rainforest. However, this reordering does not affect the optimum of a binary ranking metric.

In our evaluation protocol, ground-truth targets are binarized: cells where species $y$ was observed have relevance $y_{x,y} = 1$ (habitat, $\alpha_y(x) > 0$), while unrecorded cells have relevance $y_{x,y} = 0$ (non-habitat, $\alpha_y(x) = 0$). For any non-habitat cell, $\alpha_y(x_{\text{non}}) = 0 \implies \sigma(f_y(x_{\text{non}})) = 0$. For any true habitat cell, $\alpha_y(x_{\text{hab}}) > 0 \implies \sigma(f_y(x_{\text{hab}})) > 0$. Therefore, for any habitat site $x_{\text{hab}}$ and non-habitat site $x_{\text{non}}$:
\begin{equation}
    f_y(x_{\text{hab}}) > f_y(x_{\text{non}}).
\end{equation}
Because all observed habitat cells share identical binary relevance ($1$), any reordering among habitat locations leaves the ideal discounted cumulative gain ($\text{IDCG}$) and the empirical gain unchanged whenever top ranks are occupied by true presences. The community richness denominator $R(x)$ can never invert a habitat cell below an unsuitable background cell at the Bayes optimum, ensuring that the metric's global maximum ($\text{NDCG}@K = 1.0$) is strictly preserved.

\end{document}